\documentclass[10pt,twocolumn,letterpaper]{article}

\usepackage{cvpr}              

\usepackage{pifont}

\newcommand\blfootnote[1]{%
  \begingroup
  \renewcommand\thefootnote{}\footnote{#1}%
  \addtocounter{footnote}{-1}%
  \endgroup
}     

\usepackage{makecell}

\definecolor{cvprblue}{rgb}{0.21,0.49,0.74}
\usepackage[pagebackref,breaklinks,colorlinks,allcolors=cvprblue]{hyperref}

\usepackage{listings}
\usepackage{algorithm}
\usepackage{algpseudocode}
\usepackage{pythonhighlight}
\usepackage{wrapfig,lipsum,booktabs}
\usepackage{comment}
\usepackage[normalem]{ulem}
\usepackage{multirow}

\newcommand{\mo}[1]{{\sout{}}}

\def\paperID{14675} 
\def\confName{CVPR}
\def\confYear{2025}

\title{Faster Parameter-Efficient Tuning with Token Redundancy Reduction}

\author{Kwonyoung Kim$^1$ \quad Jungin Park$^{1*}$ \quad Jin Kim$^1$ \quad Hyeongjun Kwon$^1$ \quad Kwanghoon Sohn$^{1, 2*}$\\
$^1$Yonsei University, \quad $^2$Korea Institute of Science and Technology (KIST)\\
{\tt\small \{kyk12, newrun, kimjin928, kwonjunn01 khsohn\}@yonsei.ac.kr}
}

\begin{document}
\maketitle

\begin{abstract}
    \blfootnote{\hskip -0.2in $*$ Corresponding authors.}
    Parameter-efficient tuning (PET) aims to transfer pre-trained foundation models to downstream tasks by learning a small number of parameters.
    Compared to traditional fine-tuning, which updates the entire model, PET significantly reduces storage and transfer costs for each task regardless of exponentially increasing pre-trained model capacity.
    However, most PET methods inherit the inference latency of their large backbone models and often introduce additional computational overhead due to additional modules (e.g. adapters), limiting their practicality for compute-intensive applications.
    In this paper, we propose Faster Parameter-Efficient Tuning (FPET), a novel approach that enhances inference speed and training efficiency while maintaining high storage efficiency.
    Specifically, we introduce a plug-and-play token redundancy reduction module delicately designed for PET.
    This module refines tokens from the self-attention layer using an adapter to learn the accurate similarity between tokens and cuts off the tokens through a fully-differentiable token merging strategy, which uses a straight-through estimator for optimal token reduction.
    Experimental results prove that our FPET achieves faster inference and higher memory efficiency than the pre-trained backbone while keeping competitive performance on par with state-of-the-art PET methods.
    The code is available at \url{https://github.com/kyk120/fpet}.
    
\end{abstract}    
\section{Introduction}
\label{sec:intro}

Pre-training on large-scale web-collected data followed by fine-tuning on specific downstream tasks is a foundational paradigm that leads to state-of-the-art performance across various vision-related tasks. 
Nonetheless, modifying all parameters for each distinct task is inefficient in terms of storage, as it requires updating and storing the whole parameters for every individual task.
To address this issue, parameter-efficient tuning (PET) approaches~\cite{houlsby2019parameter, chen2022adaptformer, jia2022visual, tu2023visual, zhao2023revisit, zaken2021bitfit, jie2023fact, lian2022scaling, hu2021lora, pan2022st, li2023prefix, jie2023revisiting, zhang2022neural, karimi2021compacter, park2023dual, he2022sparseadapter, he2023parameter, pfeiffer2020adapterfusion, han2024straightforward, zhang2024parameter} have proposed to utilize a minimal number of parameters to transfer pre-trained models to downstream tasks.
They have demonstrated significant storage efficiency while attaining comparable or even surpassing performance to full-tuning. For example, \cite{jie2023revisiting} outperforms full-tuning performance on VTAB-1K~\cite{zhai2019visual} by 8\% with only less than 2\% trainable parameters.

\begin{figure}[!t]
     \centering
     \begin{subfigure}[b]{\linewidth}
         \centering
         \includegraphics[width=\linewidth]{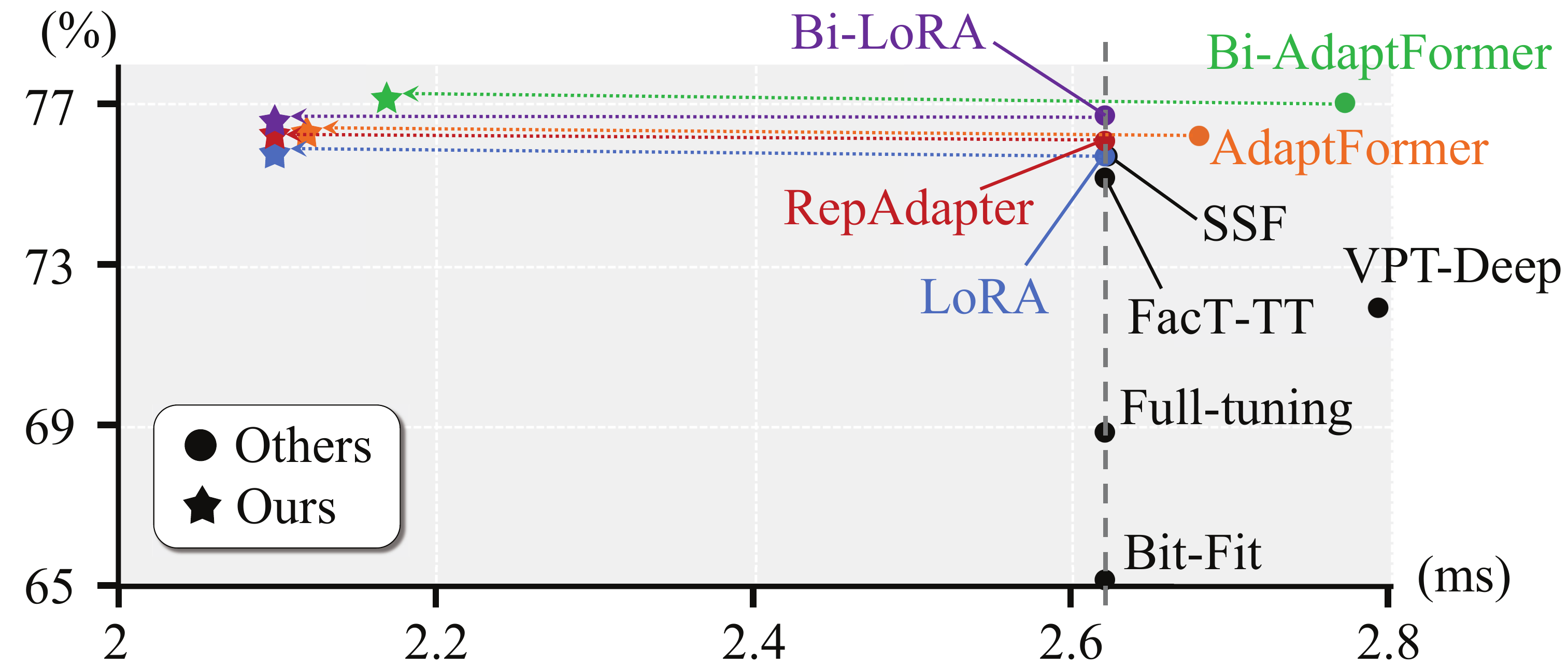}
         \caption{Accuracy (\%) vs Inference Time (ms)}
         \label{fig:fig1_graphs_a}
     \end{subfigure}
     \hfill
     \begin{subfigure}[b]{\linewidth}
         \centering
         \includegraphics[width=\linewidth]{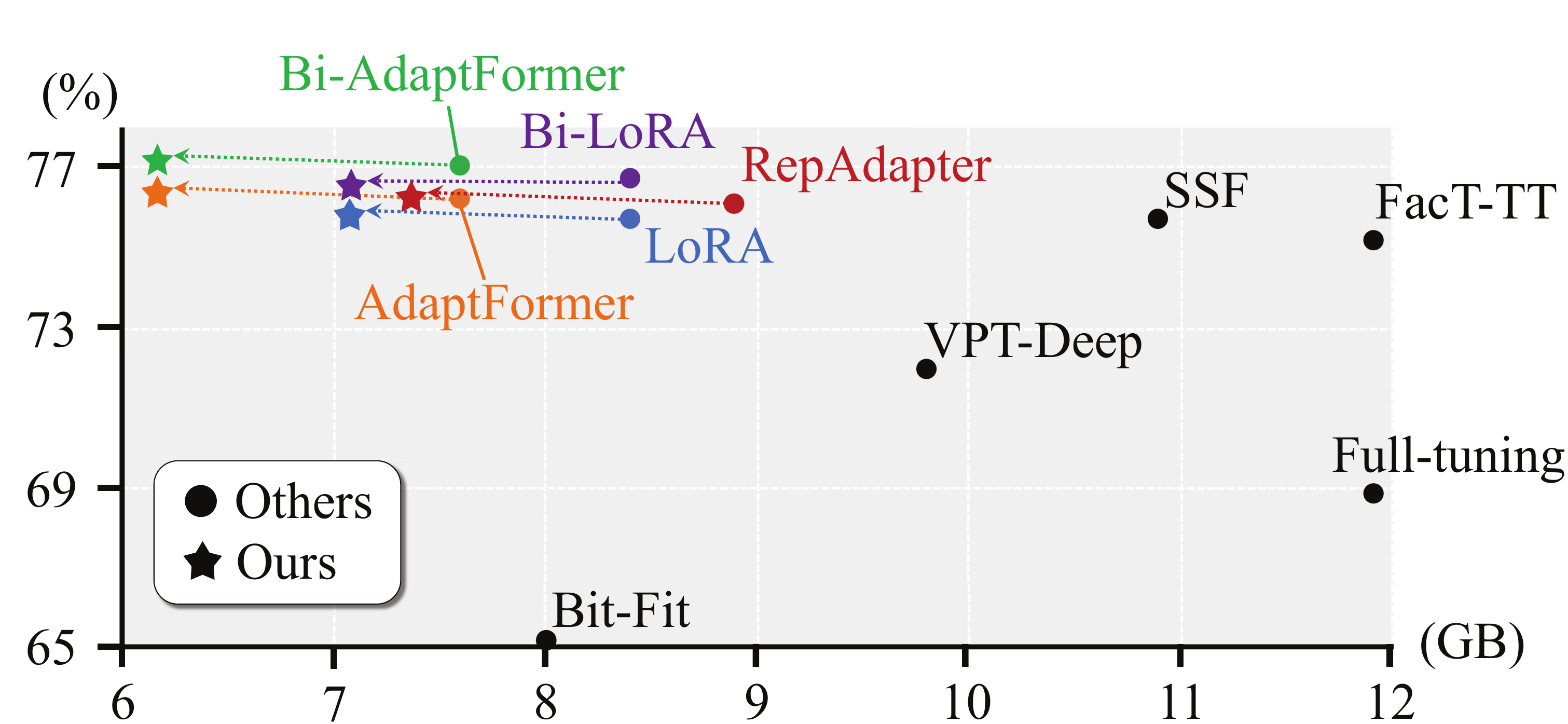}
         \caption{Accuracy (\%) vs GPU Memory usage (GB)}
         \label{fig:fig1_graphs_b}
     \end{subfigure}
     \caption{Average accuracy vs. inference time and GPU memory usage on VTAB-1K~\cite{zhai2019visual}. Our FPET significantly surpasses all existing PET methods in terms of inference speed and computation cost. Note that the dotted line in (a) represents the \textit{no inference latency} established in prior studies.}
    \label{fig:fig1_graphs}
    \vspace{-10pt}
\end{figure}

    PET methods have continually evolved with the fundamental objective of conserving resources by redundancy reduction~\cite{jie2023revisiting}.
    Specifically, they have accomplished redundancy reduction through rank decomposition with additional lightweight learnable modules (e.g. adapter)~\cite{houlsby2019parameter, hu2021lora, chen2022adaptformer, jie2023fact, jie2023revisiting, pan2022st, zhao2023revisit, zhang2022neural, karimi2021compacter, he2022sparseadapter, he2023parameter, pfeiffer2020adapterfusion}, task-specific parameter reduction with additional learnable tokens~\cite{jia2022visual, tu2023visual, li2023prefix, zhang2023towards}, or precision redundancy reduction with adapter quantization~\cite{jie2023revisiting}.
    While these strategies have made significant advances in storage efficiency, they are still insufficient to apply to real-world applications (\eg web platforms utilized by millions of users or edge devices that are constrained by resources) due to two main challenges: \textit{inference latency} and \textit{computing memory usage}.
    Most existing works often require higher inference latency and computation than full-tuning models during inference due to additional modules or parameters.
    Some works~\cite{hu2021lora, jie2023fact, lian2022scaling, zaken2021bitfit, luo2023towards} have facilitated \textit{no inference latency} compared to the backbone models, however, they inherit the original inference latency and computation requirement from the backbone.

    To address the above limitations, some approaches~\cite{han2024straightforward, zhang2024parameter} have proposed to truncate the network architecture~\cite{han2024straightforward} or disentangle task-specific and pre-trained task-agnostic knowledge~\cite{zhang2024parameter}.
    However, they still suffer from an increased number of learnable parameters and a slower inference speed~\cite{zhang2024parameter}, and an additional pre-training step~\cite{han2024straightforward}.
    Furthermore, they often exhibit degraded performance compared to state-of-the-art PET methods.

    In this paper, we explore a straightforward approach to achieve both a feasible inference latency and a training computational efficiency.
    Motivated by the recent token reduction approaches~\cite{bolya2022token, bonnaerens2023learned}, which reduces model's input space, we present Faster Parameter-Efficient Tuning (FPET) that formulates PET with token redundancy reduction.
    Instead of directly utilizing the previous methods, which provides a sub-optimal solution due to their non-differentiable nature, we introduce a fully differentiable token redundancy reduction module.
    Specifically, we incorporate a straight-through estimator (STE)~\cite{bengio2013estimating} into token reduction to make the token selection process fully differentiable.
    In addition, our FPET performs token reduction once in the backbone model's intermediate layer contrary to \cite{bolya2022token, bonnaerens2023learned}, which performs token merging based on token similarity in the early layer where the impact of the adapter is not fully manifest.
    Consequently, our FPET achieves faster inference speed, surpassing even the \textit{no inference latency} approaches~\cite{hu2021lora, jie2023fact, lian2022scaling, zaken2021bitfit, luo2023towards}, while maintaining efficiency gains during training.
    Experiments on VTAB-1K~\cite{zhai2019visual} and extensive ablation study demonstrate the effectiveness and high efficiency of our FPET, achieving a faster inference speed and a lower computation requirement than existing PET methods while attaining comparable performance to state-of-the-art.

\section{Related Work}
\label{sec:relatedwork}

\subsection{Parameter-efficient tuning}
Parameter-efficient tuning (PET) seeks to leverage large pre-trained models by tuning and storing only the minimal task-specific parameters, thereby minimizing storage requirements across a multitude of downstream tasks.

Prompt-based methods~\cite{jia2022visual, tu2023visual, li2023prefix, Kwon_2023_CVPR, Sohn_2023_CVPR, Liu_2023_CVPR, Zhu_2023_CVPR, zhang2023towards} introduce a small number of learnable tokens that are concatenated with input tokens of transformer. 
Through the self-attention layer, prompts modulate input tokens for adaptation, but the additional token count (1 to 200) results in a quadratic increase in computational complexity.

Adapter-based methods~\cite{houlsby2019parameter, chen2022adaptformer, jie2023fact, hu2021lora, pan2022st, jie2023revisiting, zhang2022neural, karimi2021compacter, park2023dual, he2022sparseadapter, he2023parameter, pfeiffer2020adapterfusion, liu2024parameterefficient}, has made significant strides in decreasing rank redundancy employing rank decomposition matrices into transformer architecture.
Especially, \cite{jie2023fact, he2023parameter, karimi2021compacter, liu2024parameterefficient} further improved the decomposition utilizing Kronecker products or butterfly factorization. 
Furthermore, \cite{jie2023revisiting} employs quantization on adapters, considerably diminishing the storage demands for each task by reducing numerical precision redundancy. 
However, their concern is limited to storage memory efficiency. 

There have been attempts to maintain inference speed without incurring any increase from adapters~\cite{hu2021lora, jie2023fact, lian2022scaling, zaken2021bitfit}. 
By employing low-rank adapters devoid of non-linear functions~\cite{hu2021lora, jie2023fact}, or utilizing a scaling and shifting module as an adapter~\cite{lian2022scaling}, they enable the integration of learned parameters into the pre-trained model in advance. Furthermore, ~\cite{zaken2021bitfit} maintains inference speed by implementing partial fine-tuning instead of additional adapters.
Despite these attempts, the improvements have been limited in maintaining the original inference speed of pre-trained models. 

Recent studies~\cite{han2024straightforward, zhang2024parameter} have proposed various approaches to enhance the efficiency of PET. These approaches include disentangled learning~\cite{zhang2024parameter} and structural reduction methods~\cite{han2024straightforward}. While these strategies have shown potential for improving efficiency, the gains are often conditional and come with trade-offs. Specifically, they may enhance training or inference efficiency but at the expense of the need for an additional pre-training step~\cite{han2024straightforward}, a significant increase in the number of learnable parameters~\cite{zhang2024parameter}, slower inference speed~\cite{zhang2024parameter} or degraded accuracy~\cite{han2024straightforward, zhang2024parameter}. These limitations can affect their applicability in practical scenarios. In contrast, our approach is designed to enhance training, inference, and parameter efficiency simultaneously, offering a more practical and balanced solution for real-world PET implementations.

\begin{figure*}[t]
     \centering
     \begin{subfigure}[b]{\textwidth}
         \centering
         \includegraphics[width=\textwidth]{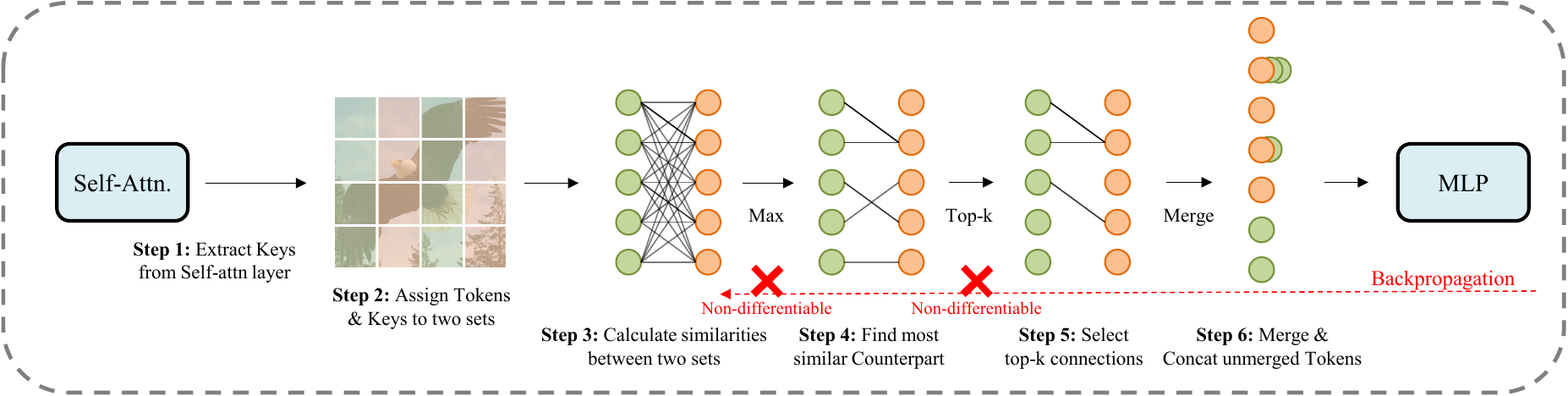}
         \caption{Bipartite soft matching~\cite{bolya2022token}}
         \label{fig:fig2_a}
     \end{subfigure}
     \begin{subfigure}[b]{\textwidth}
         \centering
         \includegraphics[width=\textwidth]{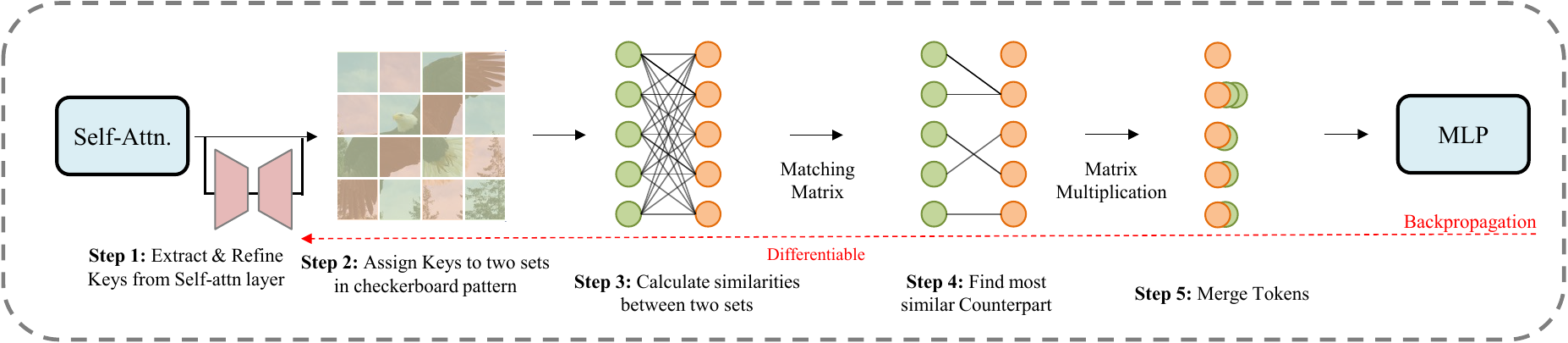}
         \caption{Bipartite differentiable matching}
         \label{fig:fig2_b}
     \end{subfigure}
     \caption{Comparison between the bipartite soft matching~\cite{bolya2022token} and the proposed bipartite differentiable matching. Our proposal is fully differentiable and refines the similarity between tokens, making token merging more optimal.}
    \label{fig:fig2}
    
\end{figure*}

\subsection{Token reduction}
Vision Transformer (ViT)~\cite{dosovitskiy2021an} has significantly advanced various vision tasks but faces challenges in resource-constrained environments due to its high computational demands. Token pruning methods methods~\cite{liang2022not,kong2022spvit,Yin_2022_CVPR,rao2021dynamicvit,yin2021adavit,liu2024a} have been developed to accelerate ViT by selectively removing less critical tokens; however, these approaches can result in permanent information loss. To address this limitation, token merging~\cite{bolya2022token} has been introduced as a technique that merges similar tokens, demonstrating improved performance over pruning methods. Unlike prior solutions such as \textit{k}-means clustering~\cite{kmeans} or graph cuts~\cite{graphcuts}, token merging is a fast, non-iterative approach. However, it cannot be optimized as it is a parameter-free heuristic method, and its bipartite soft matching process is non-differentiable due to the use of \textit{max} and top-\textit{k} operations. 
Besides, to avoid sudden changes that could negatively affect the network, this method gradually merges tokens across all layers leading to less efficiency gain in GPU memory usage.

Recent advancements using learned thresholds~\cite{bonnaerens2023learned} to merge tokens whose similarities exceed those learned thresholds.
Despite learning thresholds, they did not consider refining the similarities between tokens. Additionally, their matching process remains non-differentiable, and the dynamic reduction strategy with thresholds does not guarantee efficiency gains unless the batch size is 1.

As these methods were designed for in-domain training-free or full-tuning scenarios, we have carefully adapted token merging for PET, thereby achieving significant efficiency improvements.

\section{Preliminaries}
\subsection{Parameter-efficient tuning}
In this section, we first lay out the framework for parameter-efficient tuning. We start by concisely formulating the adapters that are commonly utilized. 

\noindent\textbf{LoRA}
\cite{hu2021lora} utilizes low-rank decomposition to approximate the variation in pre-trained parameters in response to changes in the input. Specifically, it learns the alterations in ${\mathbf{W}^q}$ and $\mathbf{W}^v$ 
within the self-attention layer to respectively modify the query and value projection weights. The adapter consist of two FC layers, $\mathbf{W}_{down} \in \mathbb{R}^{d\times h}$ and $\mathbf{W}_{up} \in \mathbb{R}^{h\times d}$ where, $d$ is token dimension and $h << d$. 
Given an input $X\in \mathbb{R}^{N\times d}$, query and value formulations are
\begin{align}\label{equ:query1}
    \begin{aligned}
    \mathbf{Q} = \mathbf{X}\mathbf{W}^q + s \cdot \mathbf{X}\mathbf{W}^{q}_{\text{down}}\mathbf{W}^{q}_{\text{up}}, \\
    \mathbf{V} = \mathbf{X}\mathbf{W}^v + s \cdot \mathbf{X}\mathbf{W}^{v}_{\text{down}}\mathbf{W}^{v}_{\text{up}},
    \end{aligned}
\end{align}
where ${\mathbf{W}^q}$ and $\mathbf{W}^v$ are frozen and $s$ is hyper-parameter.

\noindent\textbf{AdaptFormer}
\cite{chen2022adaptformer} employs low-rank decomposition as LoRA~\cite{hu2021lora} but it incorporates non-linear layer such as ReLU activation between ${\mathbf{W}_{\text{down}}}$ and $\mathbf{W}_{\text{up}}$. Additionally, AdaptFormer positions the adapter in parallel with the FFN layer in transformer layer.
Given an input $X\in \mathbb{R}^{N\times d}$, the formulation is
\begin{align}\label{equ:query2}
    \mathbf{X} = \mathbf{X} + \text{FFN}(\mathbf{X}) + s \cdot \text{ReLU}(\mathbf{X}\mathbf{W}_{\text{down}})\mathbf{W}_{\text{up}}, 
\end{align}
where, the FFN module is frozen and $s$ is a hyper-parameter.

\subsection{Bipartite soft matching}

ToMe~\cite{bolya2022token} introduced bipartite soft matching as a non-iterative, gradual method for merging tokens. The overall framework of bipartite soft matching is illustrated in \cref{fig:fig2_a}.
Given tokens $\mathbf{X}=\{ \mathbf{x}_i \} \in \mathbb{R}^{N\times d}$, where the total number of tokens is $N$ and $d$ is the feature dimension, tokens are assigned to two sets
\begin{align}\label{equ:masking1}
    \begin{aligned}
    \mathbf{X}_{\text{A}}=\{\mathbf{x}_{i} \mid i \mod 2 = 0 \} \\
    \mathbf{X}_{\text{B}}=\{\mathbf{x}_{i} \mid i \mod 2 = 1 \}
    \end{aligned}
\end{align}
To determine the optimal similarity and reduce complexity, the keys ($\mathbf{K}$) from the self-attention layer are utilized instead of token features. Keys corresponding to each token from self-attention layer, $\mathbf{K}=\{ \mathbf{k}_i \} \in \mathbb{R}^{N\times d'}$ are assigned to two sets 
\begin{align}\label{equ:masking2}
    \begin{aligned}
    \mathbf{K}_{\text{A}}=\{\mathbf{k}_{i} \mid i \mod 2 = 0 \} \\
    \mathbf{K}_{\text{B}}=\{\mathbf{k}_{i} \mid i \mod 2 = 1 \} 
    \end{aligned}
\end{align}
where $d'$ is the key dimension and $k_i$ corresponds to the $x_i$. 
The similarity matrix $\mathbf{C}\in \mathbb{R}^{N_A\times N_B}$ between two sets, $\mathbf{K}_{\text{A}}\in \mathbb{R}^{N_A\times d'}$ and $\mathbf{K}_{\text{B}}\in \mathbb{R}^{N_B\times d'}$, where $N_A$ and $N_B$ are the number of tokens in $\mathbf{X}_{\text{A}}$ and $\mathbf{X}_{\text{B}}$ respectively, is calculated as 
\begin{align}\label{equ:masking3}
    \mathbf{C}= \mathbf{K}_{\text{A}} \times \mathbf{K}_{\text{B}}^{\top}
\end{align}
For each token in set $\mathbf{K}_{\text{A}}$, to find the most similar connection within $\mathbf{K}_{\text{B}}$, \textit{max} operation along the last dimension of $\mathbf{C}$ is applied to generate max values $\mathbf{C}_{max}\in \mathbb{R}^{N_A}$ and indices $\text{I}_{max}\in \mathbb{R}^{N_A}$. 
\begin{align}\label{equ:max}
    \mathbf{C}_{max}, \, \text{I}_{max} = max(\mathbf{C}, \, dim=-1)
\end{align}
$\text{I}_{max}$ contains for each token in $\text{A}$, the most similar counterpart in $\text{B}$ and $\mathbf{C}_{max}$ contains the corresponding maximum similarity.
To match most similar \textit{k} pairs among those, the top-\textit{k} indices are extracted.
\begin{align}\label{equ:topk}
    \text{I}_{topk} = \text{top-}k(argsort(\mathbf{C}_{max}))
\end{align}
Then, $\text{I}_{max}$ and $\text{I}_{topk}$ specify which tokens in $\text{A}$ should be merged with which token in $\text{B}$. The merged tokens $\mathbf{X}_{\text{merged}}$ can be expressed as
\begin{align}\label{equ:scatter}
    \mathbf{X}_{\text{merged}} = merge(\mathbf{X}_{\text{A}}, \mathbf{X}_{\text{B}}, \text{I}_{max}, \text{I}_{topk})
\end{align}
where the \textit{merge} is a function that merges tokens by averaging matched tokens according to the input indices. 
Finally, the merged tokens are concatenated with the remaining unmerged tokens and returned as the output of the token merging module.
Note that due to the non-differentiable nature of their \textit{max} and top-\textit{k} operations, the matching process cannot be explicitly incorporated into the training objective.

\section{Faster parameter-efficient tuning}
We introduce token redundancy reduction into PET to surpass the original speed of backbone models and lower computational costs without compromising on accuracy. To achieve this goal, we have investigated optimal configuration of token redundancy reduction module suitable for faster PET.

\subsection{Token redundancy reduction module for FPET}
The recently proposed token merging method, ToMe~\cite{bolya2022token}, presents a straightforward token reduction strategy that mitigates concerns regarding information loss associated with token pruning. Rather than discarding tokens, ToMe utilizes a bipartite soft matching strategy to merge similar tokens, thereby preserving more information. To minimize information loss in our approach, we incorporate token merging as our chosen token reduction strategy.

ToMe~\cite{bolya2022token}, originally designed for in-domain training-free or full fine-tuning settings, measures similarities between tokens and gradually merges them across layers to prevent abrupt changes and minimize risk. In contrast, our approach implements the token merging module solely at the middle layer, merging half of the tokens. This modification addresses the potential for sub-optimal merging outcomes in the original ToMe methodology in the context of PET, where early-stage merging based on a similarity matrix may not fully reflect the impact of adapters. Unlike ToMe~\cite{bolya2022token}, which conducts layer-by-layer merging of 8 tokens per layer (totaling 96 tokens), our approach merges 98 tokens at the 6th layer of a 12-layer ViT-B/16.
By implementing the merging module only once at the middle layer, where token redundancy is sufficiently high, we aim to enhance accuracy while also conserving computation memory by avoiding the repeated computation of the similarity matrix across layers.

\subsection{Bipartite differentiable matching}

To further enhance our method, we reformulate the matching process to be differentiable for optimization, as depicted in \cref{fig:fig2_b}.
We begin by defining the token division sets, $\mathbf{X}_{\text{A}}$ and $\mathbf{X}_{\text{B}}$, based on observations in \cref{fig:fig2} which indicate that a na\"{i}ve bipartite soft matching split results in vertical stripe patterns when tokens are reorganized in their original image format.
In this arrangement, tokens in $\mathbf{X}_B$ can only merge with adjacent side tokens, not with those above or below. 
To address this, we divide the keys ($\mathbf{K}$) of the tokens into two sets using a checkerboard pattern:
\begin{align}\label{equ:masking6}
\begin{aligned}
    \mathbf{K}_{\text{A}}=\{\mathbf{k}_{ij} \mid (i+j) \mod 2 = 0 \} \\
    \mathbf{K}_{\text{B}}=\{\mathbf{k}_{ij} \mid (i+j) \mod 2 = 1 \} 
\end{aligned}
\end{align}
Here, $\mathbf{k}{ij}$ denotes the two-dimensional index when tokens are reorganized into their original image format.
This configuration allows each token in $\mathbf{X}{\text{B}}$ to merge with adjacent tokens, supporting a more comprehensive merging strategy.

The bipartite soft matching~\cite{bolya2022token} uses the keys ($\mathbf{K}$) from the self-attention layer, which encapsulates the information within each token, to measure similarity. However, these keys may not be the most optimal feature for similarity.
To ensure a more optimal similarity matrix $\mathbf{C}$, we implement a differentiable matching process and refine the keys ($\mathbf{K}$) using learnable adapters to achieve a more optimized matching result.
The refined keys ($\mathbf{K}'$) are represented as:
\begin{align}\label{equ:query6}
    \mathbf{K}' = \mathbf{K} + s \cdot \text{ReLU}(\mathbf{K}\mathbf{W}_{\text{down}})\mathbf{W}_{\text{up}}
\end{align}
Using these refined keys ($\mathbf{K}'$), we compute the similarity matrix $C' \in \mathbb{R}^{N_A\times N_B}$:
\begin{align}\label{equ:masking3}
    \mathbf{C'}= \mathbf{K'}_{\text{A}} \times \mathbf{K'}_{\text{B}}^{\top}
\end{align}
Instead of using the non-differentiable \textit{max} and top-\textit{k} operations, we leverage the matching matrix $\mathbf{C}_{\text{AB}} \in \mathbb{R}^{N_A\times N_B}$ where each row is a one-hot vector indicating the closest key in $\mathbf{K}_{B}$ for each key in $\mathbf{K}_{A}$. To construct $\mathbf{C}_{\text{AB}}$, we compute the soft matching matrix $\hat{\mathbf{C}}_{\text{AB}} \in \mathbb{R}^{N_A\times N_B}$ as:
\begin{align}\label{equ:query7}
   \hat{\mathbf{C}}_{\text{AB}} = \sigma (\mathbf{C} - \bar{\mathbf{C}})
\end{align}
where $\bar{\mathbf{C}} \in \mathbb{R}^{N_A \times 1}$ represents the average of the top-1 and top-2 values along the last dimension of $\mathbf{C}$ and $\sigma$ is a sigmoid function. Notably, the top-\textit{k} operation serves only to shift values, without compromising differentiability.
For each row of $\hat{\mathbf{C}}_{\text{AB}}$, only the top-1 value exceeds 0.5 since the sigmoid function outputs values greater than 0.5 for positive inputs. Thresholding $\hat{\mathbf{C}}_{\text{AB}}$ at 0.5, we derive the hard matching matrix $\mathbf{C}_{\text{AB}}$. The $\mathbf{X}_{\text{merged}}$ is then expressed as:
\begin{align}\label{equ:query8}
   \mathbf{X}_{\text{merged}} = average( \mathbf{C}^{\top}_{\text{AB}} \times \mathbf{X}_{\text{A}} + \mathbf{X}_{\text{B}} )
\end{align}
Here, \textit{average} is a function that adaptively averages tokens based on the number of matched tokens.
Each token in $\mathbf{X}_{\text{A}}$ is combined with its closest token in $\mathbf{X}_{\text{B}}$ and each summed token is averaged by the number of its matched tokens.
Since all tokens in $\mathbf{X}_{\text{A}}$ are matched to those in $\mathbf{X}_{\text{B}}$, no unmerged tokens remain. Therefore, $\mathbf{X}_{\text{merged}}$ is returned as the output of our token merging module.

While we reformulate the merging process as a differentiable matrix calculation, the hard matching matrix $\mathbf{C}_{\text{AB}}$ is non-differentiable. To approximate gradients for $\mathbf{C}_{\text{AB}}$, we use a straight-through estimator\cite{bengio2013estimating}.
Specifically, we redefine the matching matrix as:
\begin{align}\label{equ:query8}
   \tilde{\mathbf{C}}_{\text{AB}} = \hat{\mathbf{C}}_{\text{AB}} + const(\mathbf{C}_{\text{AB}} - \hat{\mathbf{C}}_{\text{AB}})
\end{align}
where \textit{const} is a function that extracts value as a constant from the tensor. Although $\mathbf{C}_{\text{AB}}$ and $\tilde{\mathbf{C}}_{\text{AB}}$ hold identical values, $\tilde{\mathbf{C}}_{\text{AB}}$ is differentiable since gradients are propagated through $\hat{\mathbf{C}}_{\text{AB}}$. 

Since the matching process is differentiable, gradients are propagated to the learnable adapter that refines the key($\mathbf{K}$). This allows for the explicit learning and the optimization of the similarity matrix.
However, allowing further backward propagation of these gradients can negatively impact accuracy due to the resulting unnecessary push-and-pull effect among tokens, similar to what is observed in contrastive loss. While the push-and-pull effect is very beneficial in self-supervised learning for distinguishing features among different samples, however, such an effect among tokens is undesirable for our task. Therefore, we halt the further propagation of these gradients to the backbone. For more detailed understanding, our code implementation is included in the supplementary materials.

\section{Experiments}
In this section, we demonstrate the superiority of our proposed method in terms of both efficiency gain and accuracy. For our implementation, we integrate our token merging module at the 6\textit{th} layer along with a quantized adapter~\cite{zhao2023revisit} for key ($\mathbf{K}$) refinement. The training epoch is 100 using AdamW optimizer. Since our proposed method can be applied to off-the-shelf PET methods in plug-and-play manner, we evaluate our method on 5 state-of-the-art PET methods, RepAdapter~\cite{luo2023towards}, LoRA~\cite{hu2021lora}, AdaptFormer~\cite{chen2022adaptformer}, Bi-LoRA~\cite{jie2023revisiting} and Bi-AdaptFormer~\cite{jie2023revisiting}.

\subsection{Datasets}
To demonstrate the efficacy of our method, we conducted evaluations across a range of downstream tasks using VTAB-1K~\cite{zhai2019visual}, which encompasses 19 diverse image classification tasks spanning different domains. VTAB-1K is divided into three categories: \textit{Natural}, \textit{Specialized}, and \textit{Structured}. The \textit{Natural} category includes classic vision tasks with images captured by standard cameras. In contrast, the \textit{Specialized} category comprises images from natural scenes captured with specialized equipment, such as those used in satellite or medical imaging. The \textit{Structured} category is centered on understanding scene structure, involving tasks like object counting or 3D depth prediction, often with images from simulated environments. Each dataset contains 1000 samples for training and validation. Following prior works~\cite{jia2022visual, jie2023fact, jie2023revisiting, lian2022scaling, zhang2022neural}, we train models using training and validation samples, and report the top-1 accuracy on test set.

\subsection{Metrics} 
We assess the performance of our approach using several metrics: accuracy, inference time, FLOPs and GPU memory usage. Accuracy evaluations are conducted on the VTAB-1K dataset, with 'Average' referring to the mean accuracy across the three groups. GPU memory usage is recorded during the training phase with batch size 64, whereas inference time and FLOPs are measured during testing with batch size 256. The inference time metric represents the time taken to process a single image. All experiments are implemented on single NVIDIA A6000 GPUs. 

\begin{table}[t]
  \centering
  \resizebox{\linewidth}{!}{
  \begin{tabular}{@{}lcccc@{}}
    \toprule
     Method & Acc (\%) & Time (ms) & FLOPs (G) & Mem (GB) \\
    \midrule
    \multicolumn{3}{@{} l}{Traditional Fine-Tuning} \\ \midrule
    Full & 68.9 & 2.62 & 17.6 & 11.9 \\
    Linear & 57.6 & 2.62 \textcolor{black}{(+0.0\%)} & 17.6 \textcolor{black}{(+0.0\%)} & 2.7 \textcolor{red}{(-77.3\%)} \\
    \midrule
    \multicolumn{3}{@{} l}{Parameter-Efficient Tuning} \\ \midrule
    VPT-Deep~\cite{jia2022visual} 
    & 72.0 & 2.79 \textcolor{blue}{(+6.5\%)} & 18.5 \textcolor{blue}{(+5.1\%)} & 9.8 \textcolor{red}{(-26.5\%)} \\
    BitFit~\cite{zaken2021bitfit} 
    & 65.2 & 2.62 \textcolor{black}{(+0.0\%)} & 17.6 \textcolor{black}{(+0.0\%)} & 8.0 \textcolor{red}{(-33.0\%)} \\
    SSF~\cite{lian2022scaling} 
    & 75.7 & 2.62 \textcolor{black}{(+0.0\%)} & 17.6 \textcolor{black}{(+0.0\%)} & 10.9 \textcolor{red}{(-8.2\%)} \\
    FacT-TT~\cite{jie2023fact} 
    & 75.2 & 2.62 \textcolor{black}{(+0.0\%)} & 17.8 \textcolor{blue}{(+1.1\%)} & 11.9 \textcolor{black}{(-0.0\%)} \\
    
    RepAdapter~\cite{luo2023towards} 
    & 76.1 & 2.62 \textcolor{black}{(+0.0\%)} & 17.6 \textcolor{black}{(+0.0\%)} & 8.9 \textcolor{red}{(-25.6\%)} \\

    LoRA~\cite{hu2021lora} 
    & 75.7 & 2.62 \textcolor{black}{(+0.0\%)} & 17.6 \textcolor{black}{(+0.0\%)} & 8.4 \textcolor{red}{(-29.6\%)} \\
    AdaptFormer~\cite{chen2022adaptformer} 
    & 76.2 & 2.68 \textcolor{blue}{(+1.5\%)} & 17.6 \textcolor{black}{(+0.0\%)} & 7.6 \textcolor{red}{(-36.0\%)} \\
    Bi-LoRA~\cite{jie2023revisiting} 
    & 76.7 & 2.62 \textcolor{black}{(+0.0\%)} & 17.6 \textcolor{black}{(+0.0\%)} & 8.4 \textcolor{red}{(-29.4\%)} \\
    Bi-AdaptFormer~\cite{jie2023revisiting} 
    & 77.0 & 2.77 \textcolor{blue}{(+5.7\%)} & 17.7 \textcolor{blue}{(+0.6\%)} & 7.6 \textcolor{red}{(-35.9\%)}  \\ 
    
    \midrule
    \multicolumn{3}{@{} l}{Efficiency-gained Parameter-Efficient Tuning} \\ \midrule

    SynQT~\cite{zhang2024parameter} 
    & 72.9 & 2.83 \textcolor{blue}{(+8.0\%)} & 16.84 \textcolor{red}{(-4.3\%)} & 3.6 \textcolor{red}{(-69.9\%)}  \\

    Pruned RepAdapter~\cite{han2024straightforward}
    & 74.8 & [1.06, 2.62] & [7.1, 17.6] & [5.9, 8.9]  \\

    \midrule
    \multicolumn{3}{@{} l}{Faster Parameter-Efficient Tuning (Ours)} \\ \midrule
    RepAdapter 
    & 76.1 & 2.10 \textcolor{red}{(-19.8\%)} & 13.3 \textcolor{red}{(-24.4\%)} & 7.4 \textcolor{red}{(-38.0\%)} \\
    LoRA 
    & 75.6 & 2.10 \textcolor{red}{(-19.8\%)} & 13.3 \textcolor{red}{(-24.4\%)} & 7.1 \textcolor{red}{(-40.3\%)} \\
    AdaptFormer 
    & 76.2 & 2.12 \textcolor{red}{(-19.1\%)} & 13.5 \textcolor{red}{(-23.3\%)} & 6.2 \textcolor{red}{(-47.9\%)} \\
    Bi-LoRA 
    & 76.4 & 2.10 \textcolor{red}{(-19.8\%)} & 13.3 \textcolor{red}{(-24.4\%)} & 7.1 \textcolor{red}{(-40.3\%)} \\
    Bi-AdaptFormer 
    & 77.0 & 2.17 \textcolor{red}{(-17.2\%)} & 13.5 \textcolor{red}{(-23.3\%)} & 6.2 \textcolor{red}{(-47.8\%)} \\
    \bottomrule
  \end{tabular}}
  \caption{Comparison with state-of-the-art methods in terms of accuracy, inference time per image, FLOPs, and GPU memory usage during training. In terms of efficiency, we present both the absolute values and the relative gap in comparison to the full fine-tuning method.
  For Pruned RepAdapter~\cite{han2024straightforward}, the values are shown as ranges to indicate its inconsistent efficiency across datasets.
  }
  \label{tab:efficiency}
\end{table}

\begin{table*}[ht]
\centering
\resizebox{\linewidth}{!}{
\begin{tabular}{@{}lcc|ccccccc|cccc|cccccccc@{}}
\toprule
& & & \multicolumn{7}{|c}{\textbf{Natural}} & \multicolumn{4}{|c}{\textbf{Specialized}} & \multicolumn{8}{|c}{\textbf{Structured}} \\

& \rotatebox{90}{\# param (M)} & \rotatebox{90}{Average} & \rotatebox{90}{CIFAR100} & \rotatebox{90}{Caltech101} & \rotatebox{90}{DTD} & \rotatebox{90}{Flowers} & \rotatebox{90}{Pets} & \rotatebox{90}{SVHN} & \rotatebox{90}{Sun397} & \rotatebox{90}{Camelyon} & \rotatebox{90}{EuroSAT} & \rotatebox{90}{Resisc45} & \rotatebox{90}{Retinopathy} & \rotatebox{90}{Clevr-Count} & \rotatebox{90}{Clevr-Dist} & \rotatebox{90}{DMLab} & \rotatebox{90}{KITTI-Dist} & \rotatebox{90}{dSpr-Loc} & \rotatebox{90}{dSpr-Ori} & \rotatebox{90}{sNORB-Azim} & \rotatebox{90}{sNORB-Ele} \\ \midrule

\multicolumn{3}{@{} l}{Traditional Fine-Tuning} \\ \midrule

Full & 85.8 & 68.9 & 68.9 & 87.7 & 64.3 & 97.2 & 86.9 & 87.4 & 38.8 & 79.7 & 95.7 & 84.2 & 73.9 & 56.3 & 58.6 & 41.7 & 65.5 & 57.5 & 46.7 & 25.7 & 29.1 \\

Linear & 0 & 57.6 & 64.4 & 85.0 & 63.2 & 97.0 & 86.3 & 36.6 & 51.0 & 78.5 & 87.5 & 68.5 & 74.0 & 34.3 & 30.6 & 33.2 & 55.4 & 12.5 & 20.0 & 9.6 & 19.2 \\ \midrule

\multicolumn{3}{@{} l}{Parameter-Efficient Tuning} \\ \midrule

VPT-Shallow~\cite{jia2022visual} 
& 0.06 & 67.8 & \textbf{77.7} & 86.9 & 62.6 & 97.5 & 87.3 & 74.5 & 51.2 & 78.2 & 92.0 & 75.6 & 72.9 & 50.5 & 58.6 & 40.5 & 67.1 & 68.7 & 36.1 & 20.2 & 34.1 \\

VPT-DEEP~\cite{jia2022visual} 
& 0.53 & 72.0 & \textbf{78.8} & 90.8 & 65.8 & 98.0 & 88.3 & 78.1 & 49.6 & 81.8 & 96.1 & 83.4 & 68.4 & 68.5 & 60.0 & 46.5 & 72.8 & 73.6 & 47.9 & 32.9 & 37.8 \\

NOAH~\cite{zhang2022neural} 
& 0.36 & 75.5 & 69.6 & 92.7 & 70.2 & 99.1 & 90.4 & 86.1 & 53.7 & 84.4 & 95.4 & 83.9 & 75.8 & 82.8 & 68.9 & 49.9 & \textbf{81.7} & 81.8 & 48.3 & 32.8 & 44.2 \\

BitFit~\cite{zaken2021bitfit} 
& 0.10 & 65.2 & 72.8 & 87.0 & 59.2 & 97.5 & 85.3 & 59.9 & 51.4 & 78.7 & 91.6 & 72.9 & 69.8 & 61.5 & 55.6 & 32.4 & 55.9 & 66.6 & 40.0 & 15.7 & 25.1 \\

SSF~\cite{lian2022scaling} 
& 0.24 & 75.7 & 69.0 & 92.6 & \textbf{75.1} & \textbf{99.4} & \textbf{91.8} & 90.2 & 52.9 & 87.4 & 95.9 & \textbf{87.4} & 75.5 & 75.9 & 62.3 & \textbf{53.3} & 80.6 & 77.3 & \textbf{54.9} & 29.5 & 37.9 \\

FacT-TT~\cite{jie2023fact} 
& 0.04 & 75.3 & 71.3 & 89.6 & 70.7 & 98.9 & 91.0 & 87.8 & 54.6 & 85.2 & 95.5 & 83.4 & 75.7 & 82.0 & \textbf{69.0} & 49.8 & 80.0 & 79.2 & 48.4 & 34.2 & 41.4 \\

RepAdapter~\cite{luo2023towards} 
& 0.22 & 76.1 & 72.4 & 91.6 & 71.0 & 99.2 & 91.4 & \textbf{90.7} & 55.1 & 85.3 & 95.9 & 84.6 & 75.9 & 82.3 & 68.0 & 50.4 & 79.9 & 80.4 & 49.2 & \textbf{38.6} & 41.0 \\

LoRA~\cite{hu2021lora} 
& 0.29 & 75.7 & 70.9 & \textbf{93.0} & 70.0 & 99.1 & 91.2 & 86.0 & 55.8 & 87.1 & 94.6 & 82.4 & 74.7 & 83.5 & 63.1 & 51.7 & 79.3 & 85.2 & 51.8 & 33.5 & 43.6 \\

AdaptFormer~\cite{chen2022adaptformer} 
& 0.16 & 76.2 & 72.0 & 92.7 & 70.2 & 99.3 & 91.0 & 87.5 & 54.8 & 87.4 & 95.2 & 85.2 & 75.2 & 83.6 & 62.6 & 52.1 & 81.0 & 86.2 & 53.1 & 34.5 & 40.3 \\

Bi-LoRA~\cite{jie2023revisiting} 
& 1.17 & 76.7 & 72.1 & 91.7 & 71.2 & 99.1 & 91.4 & 90.2 & 55.8 & 87.0 & 95.4 & 85.5 & 75.5 & 83.1 & 64.1 & 52.2 & 81.3 & \textbf{86.4} & 53.5 & 36.7 & \textbf{44.4} \\

Bi-AdaptFormer~\cite{jie2023revisiting}  
& 0.64 & \textbf{77.0} & 74.1 & 92.4 & 72.1 & 99.3 & \textbf{91.6} & 89.0 & \textbf{56.3} & \textbf{88.2} & 95.2 & 86.0 & \textbf{76.2} & \textbf{83.9} & 63.6 & \textbf{53.0} & 81.4 & 86.2 & \textbf{54.8} & 35.2 & 41.3 \\ 

\midrule
\multicolumn{3}{@{} l}{Efficiency-gained Parameter-Efficient Tuning} \\ 
\midrule

SynQT~\cite{zhang2024parameter} 
& 2.72 & 72.9 & 70.9 & 89.7 & 68.8 & 98.5 & 89.6 & 77.8 & 50.6 & 82.3 & \textbf{96.7} & 83.5 & 75.2 & 71.8 & 62.7 & 48.5 & 75.4 & 74.1 & 49.0 & 31.7 & 36.1 \\

Pruned RepAdapter~\cite{han2024straightforward}
& 0.18 & 74.8 & 71.4 & 87.3 & 68.1 & 96.0 & 89.9 & 89.3 & 53.4 & 85.0 & 95.3 & 81.9 & 75.2 & 80.9 & \textbf{69.8} & 50.5 & 80.7 & 80.5 & 47.1 & 35.7 & 41.0 \\

\midrule
\multicolumn{3}{@{} l}{Faster Parameter-Efficient Tuning (Ours)} \\ 
\midrule

RepAdapter
& 0.23 & 76.1 & 72.1 & 91.5 & 71.8 & 99.3 & 90.7 & \textbf{90.3} & 55.0 & 85.2 & \textbf{96.2} & 84.5 & 75.6 & 82.2 & 67.7 & 49.7 & 79.9 & 82.2 & 48.7 & \textbf{36.9} & 41.7 \\

LoRA 
& 0.30 & 75.6 & 70.1 & 92.7 & 69.4 & 99.1 & 90.8 & 85.4 & 55.6 & 87.2 & 94.6 & 82.5 & 74.1 & 83.0 & 63.4 & 50.6 & \textbf{81.6} & 84.7 & 51.5 & 34.3 & 43.3\\

AdaptFormer 
& 0.17 & 76.2 & 71.3 & \textbf{93.5} & 69.9 & 99.3 & 90.7 & 87.0 & 54.7 & 87.5 & 95.1 & 84.5 & \textbf{76.2} & 83.6 & 63.1 & 52.2 & 81.3 & \textbf{87.1} & 54.1 & 33.5 & 40.2 \\

Bi-LoRA 
& 1.18 & 76.4 & 71.9 & 91.1 & 70.9 & 99.1 & 90.5 & 89.4 & 55.9 & 87.4 & 94.7 & 84.4 & 74.9 & 83.5 & 65.1 & 52.1 & 79.7 & 85.8 & 54.2 & 36.7 & \textbf{44.4}  \\

Bi-Adaptformer 
& 0.64 & \textbf{77.0} & 74.1 & 92.8 & \textbf{72.5} & \textbf{99.4} & 91.1 & 89.6 & \textbf{56.2} & \textbf{88.3} & 94.9 & \textbf{86.3} & 75.3 & \textbf{83.8} & 63.0 & 52.8 & 81.4 & 85.7 & 54.4 & 35.9 & 42.2  \\

\bottomrule
\end{tabular}}
\caption{Comparison with state-of-the-art methods on VTAB-1K~\cite{zhai2019visual} benchmark. Average indicates average accuracy over three groups. \# param denotes the number of learnable parameters.}
\label{tab:sota}
\end{table*}

\subsection{Comparison to the state-of-the-art methods}
We evaluate our method against various state-of-the-art approaches on VTAB-1K, which encompasses a diverse range of downstream datasets. In this subsection, we provide detailed comparisons of PET methods and efficiency-gained PET methods, respectively.
\subsubsection{PET methods.}
In this section, we compare our method with full fine-tuning, linear probing which trains classification head only, VPT~\cite{jia2022visual}, NOAH~\cite{zhang2022neural}, LoRA~\cite{hu2021lora}, SSF~\cite{lian2022scaling}, AdaptFormer~\cite{chen2022adaptformer}, BitFit~\cite{zaken2021bitfit}, FacT-TT~\cite{jie2023fact}, Bi-LoRA and Bi-Adaptformer~\cite{jie2023revisiting}.
All baseline models utilize the ViT-B/16, pre-trained on ImageNet-21K in a supervised fashion, as their backbone. For LoRA~\cite{hu2021lora} and AdaptFormer~\cite{chen2022adaptformer}, we set the hidden dimension to 8. The settings for other baseline models follow the configurations reported in their respective original papers. 

As shown in \cref{tab:efficiency} and \cref{fig:fig1_graphs}, our method demonstrates constant efficiency gain for both training and inference. Remarkably, we achieve faster inference speeds compared to PET methods, including traditional fine-tuning approaches that do not necessitate additional computation during inference. 
While certain PET methods~\cite{hu2021lora, jie2023fact, lian2022scaling, zaken2021bitfit} have been developed with the goal of achieving no-latency inference, they have only succeeded in matching the speed of traditional fine-tuning methods, leaving inherent limitations.
Our method, however, overcomes these constraints with a 19.8\% increase in speed over the original pre-trained models, demonstrating the superiority of our approach. In terms of GPU memory consumption, we achieve a 40\% reduction compared to full fine-tuning. While linear probing exhibits greater memory efficiency owing to its simplicity, it significantly underperforms in accuracy, inference speed, and FLOPs compared to our method.

The higher FLOPs observed in VPT-Deep~\cite{jia2022visual} compared to other methods suggest that an increase in the number of tokens leads to a quadratic rise in computational complexity. In contrast, our method succeed in achieving an average reduction of 24\% in FLOPs by reducing token redundancy, compared to each original implementation.
Our method attains efficiency gains with minimal impact on accuracy. Notably, when applied to Bi-AdaptFormer~\cite{jie2023revisiting}, compared to the original implementation, it results in negligible accuracy loss while achieving improvements of 21.7\%, 23.7\% and 18.4\% in inference time, FLOPs and GPU memory usage respectively.

\cref{tab:sota} presents the comprehensive results of our comparison with state-of-the-art models. Through our framework, while realizing efficiency gains, models still retain their superiority in both accuracy and parameter efficiency. As we employ only a very lightweight adapter to refine the similarity matrix, less than 0.005M, the increase in the number of trainable parameters is minimal.

\subsubsection{Efficiency-gained PET methods.}
We further compare our method with efficiency-gained PET methods including SynQT~\cite{zhang2024parameter} and Pruned RepAdapter~\cite{han2024straightforward}. As shown in \cref{tab:efficiency}, while SynQT significantly reduces GPU memory usage during training, it introduces a much slower inference speed compared to the backbone model, and its FLOPs reduction is much smaller than ours. Additionally, as detailed in \cref{tab:sota}, SynQT employs 2.3$\times$ to 16$\times$ more learnable parameters than our approach, resulting in lower accuracy. 

Pruned RepAdapter~\cite{han2024straightforward} exhibits varying levels of efficiency gains depending on the dataset, as it adaptively determines the number of layers to prune during the pre-training stage for each dataset. In \cref{tab:efficiency}, the maximum and minimum efficiency gain is presented. For the EuroSAT~\cite{helber2019eurosat} dataset, only 5 layers are used leading to a 59\%, 57\%, and 50\% reduction in inference time, FLOPs, and GPU memory usage respectively. However, for datasets such as dspr-Ori~\cite{higgins2016beta} and sNORB-Ele~\cite{smallnorb}, all 12 layers are utilized, resulting in efficiency metrics nearly identical to those of the original RepAdapter~\cite{luo2023towards}. Although Pruned RepAdapter~\cite{han2024straightforward} achieves average efficiency gains, its full-layer implementation is not suitable for resource-constrained environments making real-world applications challenging. In contrast, our method consistently achieves efficiency gains across different datasets, offering a more practical and feasible solution for real-world scenarios.

\subsection{Ablation study}
In this section, we present ablation studies to further investigate specific efficacy of our proposed methodology.

\subsubsection{Comparison to other token reduction method.}
As shown in \cref{tab:tokenreduction}, our proposed method achieves superior efficiency gains compared to existing token reduction techniques. Unlike ToMe~\cite{bolya2022token}, which merges a fixed number of tokens at every layer, LTMP~\cite{bonnaerens2023learned} dynamically determines the number of tokens to prune using a learnable threshold. To enable gradient flow, LTMP masks rather than reduces the number of tokens, requiring a modified attention module with masked softmax, which incurs memory overhead as outlined in \cref{tab:tokenreduction}. Moreover, LTMP does not yield inference-time efficiency gains when the batch size exceeds one. While ToMe improves inference speed, it fails to reduce GPU memory usage due to bipartite soft matching across all layers. Both ToMe and LTMP incur larger accuracy drops than our method, highlighting the effectiveness of our design in balancing performance and efficiency within the PET framework.

\subsubsection{Effectiveness of each component.}
In \cref{tab:tokenreduction}, we further evaluate the effectiveness of each component through progressive integration. Applying bipartite soft matching (BSM)\cite{bolya2022token} at a middle layer improves both accuracy and efficiency over the original ToMe\cite{bolya2022token} implementation. Incorporating the checkerboard pattern yields a slight accuracy gain, demonstrating the benefit of spatial priors. Adding the key refinement adapter without gradient stopping, however, significantly degrades accuracy due to interference in token representations, inducing unnecessary push and pull effect among tokens. Lastly, applying gradient stopping leads to our best overall performance, achieving strong accuracy while maintaining efficiency.

\subsubsection{Trade-offs between efficiency and accuracy.}
\begin{figure}[!t]
     \centering
     \includegraphics[width=\linewidth]{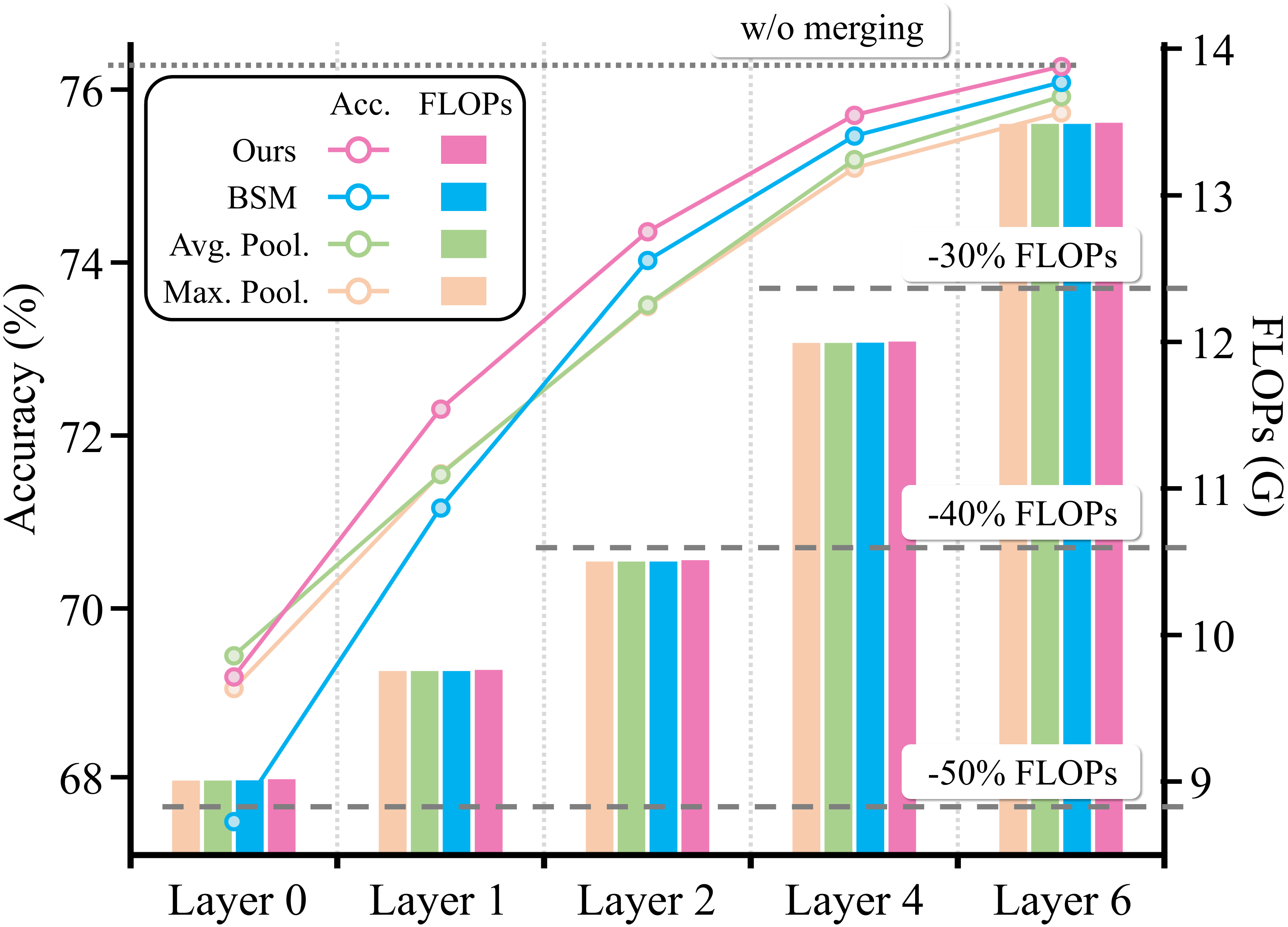}
     \hfill
     \vspace{-10pt}
     \caption{Trade-off between FLOPs (right y-axis, bar plot) and accuracy (left y-axis, line plot). All models are ViT-B/16 consisting of 12 transformer layers with AdaptFormer~\cite{chen2022adaptformer}. BSM refers to bipartite soft matching~\cite{bolya2022token}. All methods reduce 196 tokens to 98 tokens at different layers.}
    \label{fig:layer}
\end{figure}

In \cref{fig:layer}, we present the trade-off between efficiency and accuracy when applying our token reduction module at different transformer layers. The module can be flexibly applied at various layers to balance computational cost and model performance. When applied at layer 6, our method achieves the upper-bound accuracy of 76.22\%, equivalent to the original PET method without token merging, while reducing FLOPs by 23.3\%. At layer 4, a slight accuracy drop of 0.57\% yields a 31.82\% FLOPs reduction, and at layer 2, a 1.93\% drop corresponds to a 40.34\% reduction. Despite these trade-offs, our method remains competitive with both the original and efficiency-gained PET methods.

We also compare our method with other token merging strategies, including bipartite soft matching (BSM)~\cite{bolya2022token}, average pooling, and max pooling. All methods except max pooling merge tokens by averaging matched pairs. Average pooling relies on deterministic matching, BSM uses a heuristic and non-differentiable approach, while our method employs a learnable bipartite matching technique. Compared to others, our method achieves higher accuracy with negligible FLOPs overhead. At layer 0, where tokens are minimally processed and adapters are not yet applied, the regional prior is competitive to similarity-based methods. From layer 1 onward, our method consistently outperforms naive pooling, demonstrating more optimal token reduction. Notably, BSM shows degraded accuracy particularly in early layers, underscoring robustness and practicality of our method. Detailed numerical results corresponding to \cref{fig:layer} are provided in the supplementary materials.

\begin{table}[t] 
  \centering
  \resizebox{\linewidth}{!}{
  \begin{tabular}{@{}lccc@{}}
    \toprule
     Method & $\Delta$Acc (\%) & $\Delta$Time (ms) & $\Delta$Mem (GB) \\
    \midrule
    w/o merging & 0 & 0 & 0\\
    LTMP~\cite{bonnaerens2023learned} & -0.38 & +0.01 & +5.84 \\
    ToMe~\cite{bolya2022token} & -0.27 & -0.56 & -0.21 \\
    \midrule
    BSM~\cite{bolya2022token} at 6\textit{th} layer & -0.18 & -0.61 & -1.38 \\
    + checkerboard & -0.17 & -0.61 & -1.38 \\
    \makecell[l]{+ key refinement \\ (w/o gradient stopping)} & -1.70 & -0.61 & -1.41 \\
    \midrule
    Ours & \textbf{-0.05} & \textbf{-0.62} & \textbf{-1.41} \\
    \bottomrule
  \end{tabular}
  }
  \caption{Performance on other token reduction strategies. All models are ViT-B/16 consist of 12 transformer layers with Bi-AdaptFormer~\cite{jie2023revisiting}. Each model starts with 197 tokens. LTMP reduces a variable number of tokens, while ToMe merges 8 tokens at each of the 12 layers. For BSM at 6\textit{th} layer and the methods below, 98 tokens are merged at the 6\textit{th} layer only. All methods are applied during both training and inference time.}

  \label{tab:tokenreduction}
  
 \end{table}

\begin{table}[t] 
  \centering
  \resizebox{\linewidth}{!}{
  \begin{tabular}{@{}lcccc@{}}
    \toprule
     Method & AdaptFormer~\cite{chen2022adaptformer} & FPET-AdaptFormer & LoRA~\cite{hu2021lora} & FPET-LoRA \\
    \midrule
    Training time (s/it) & 5.62 & 4.52 \textcolor{red}{(-19.57\%)} & 6.24 & 5.03 \textcolor{red}{(-19.39\%)}\\
   
    \bottomrule
  \end{tabular}
  }
  \caption{Average training time across VTAB-1K~\cite{zhai2019visual} datasets.}

  \label{tab:training_time}
  
 \end{table}

\subsection{Further analysis}
In this section, we further highlight the superiority of FPET. As shown in \cref{tab:training_time}, FPET significantly improves training efficiency, reducing training time by 19\% compared to the original implementation. The latency introduced by our module is also negligible, accounting for only 0.56\% of the total pipeline latency. We evaluate FPET on other backbones, including DeiT-S and ViT-L, as reported in \cref{tab:otherbackbone}. Further results covering additional backbones, few-shot learning on the FGVC dataset, cross-modal retrieval and other analysis are provided in the supplementary materials, further demonstrating the robustness and versatility of our approach. Visualizations of the merged tokens are also included for qualitative analysis.

\begin{table} 
  \centering
  \resizebox{\linewidth}{!}{
  \begin{tabular}{@{}lcclll@{}}
    \toprule
     Model & Method & Acc (\%) & Time (ms) & FLOPs (G) & Mem (GB) \\
    
    \midrule
    \multirow{2}{*}{ViT-L} 
    & LoRA~\cite{hu2021lora} & 76.0 & 8.67 & 61.8 & 19.6\\
    & FPET-LoRA & 76.0 & 6.60 \textcolor{red}{(-23.9\%)} & 46.7 \textcolor{red}{(-24.4\%)} & 15.5 \textcolor{red}{(-21.0\%)} \\
    
    \midrule
    \multirow{2}{*}{DeiT-S} 
    &AdaptFormer\cite{chen2022adaptformer} & 70.3 & 1.13 & 4.8 & 4.4 \\
    &FPET-AdaptFormer & 70.3 & 0.94\textcolor{red}{(-20.7\%)} & 3.7\textcolor{red}{(-23.6\%)} & 3.8\textcolor{red}{(-12.5\%)} \\

    \bottomrule
  \end{tabular}}
  \caption{Model performance on other backbones.}
  \label{tab:otherbackbone}
\end{table}

\section{Conclusion}
    In this paper, we extend the concept of parameter efficiency in parameter-efficient tuning (PET) by exploring both inference latency and training computational efficiencyto enhance the applicability of PET. We propose Faster Parameter-Efficient Tuning (FPET), a novel framework that formulates PET as a token redundancy reduction problem. Our approach formulates token reduction in a parameter-efficient and fully differentiable manner, enabling optimal token reduction for PET.
    Our FPET significantly improves inference speed and training efficiency while attaining comparable performance to the previous PET methods, demonstrating the effectiveness of the proposed token redundancy reduction module. As our FPET can be seamlessly integrated with existing PET techniques, we hope our study will foster research and provide a foundation of practical PET for real-world applications.

\section{Acknowledgement}
This work was supported by the National Research Foundation of Korea(NRF) grant funded by the Korea government(MSIT) (RS-2025-00515741) and the Yonsei Signature Research Cluster Program of 2024 (2024-22-0161).

{
    \small
    \bibliographystyle{ieeenat_fullname}
    \bibliography{main}
}

\clearpage
\newpage
\appendix
\setcounter{table}{0}
\renewcommand{\thetable}{A\arabic{table}}
\setcounter{figure}{0}
\renewcommand{\thefigure}{A\arabic{figure}}

\section*{Appendix}
\blfootnote{\hskip -0.2in $*$ Corresponding authors.}
In this supplementary material, we provide additional experiment results including visualizations of merging outcomes and the performance evaluations on few-shot learning tasks, cross-modal retrieval tasks, full datasets, and other backbones to demonstrate the superiority of FPET across diverse settings. We provide pseudo-code implementation and complete source code files to offer a more in-depth understanding of our proposed method. Finally, for clarity, we detail hyper-parameters for our experiments and dataset configurations of the datasets we utilized.

\section{Additional Analysis}

\subsection{Performance on other backbones}
We extend the evaluation of our method to include other backbones, such as ViT-S, ViT-L, DeiT-S and DeiT-B, to assess its generalizability across different model architectures. ViT-S and DeiT-S features a token dimension that is half the size of ViT-B. In contrast, ViT-L employs a larger token dimension and consists of 24 transformer layers which is twice deeper layers than ViT-B, signifying a more complex model structure. DeiT-B shares a similar structure with ViT-B but utilizes additional distillation token. In all cases, our method is implemented at the middle layer. Our results, as detailed in \cref{tab:otherbackbone}, demonstrate that our approach not only maintains very competitive accuracy in comparison to the original implementations but also achieves significant efficiency gains, underscoring the generalizability of our method across different backbone architectures.

\subsection{Comparison with rank redundancy reduction}
We compare rank redundancy reduction with token reduction by benchmarking our method against a model with a hidden dimension of 1. As shown in \cref{tab:rank}, token reduction demonstrates superior performance in both accuracy and efficiency.
Despite a significant reduction in the hidden dimension, the gains in efficiency are limited, while there is a notable decrease in performance. In contrast, token reduction not only offers greater efficiency improvements but also maintains accuracy. Therefore, in the context of enhancing efficiency, reducing token redundancy emerges as a significantly more effective strategy than reducing rank.

\begin{table} 
  \centering
  \resizebox{\linewidth}{!}{
  \begin{tabular}{@{}lcclll@{}}
    \toprule
     Model & Method & Acc (\%) & Time (ms) & FLOPs (G) & Mem (GB) \\
    \midrule

    \multirow{2}{*}{ViT-S} 
    & Bi-LoRA~\cite{jie2023revisiting} & 74.9 & 0.67 & 4.7 & 5.0\\
    & FPET-Bi-LoRA & 74.9 & 0.59 \textcolor{red}{(-15.8\%)} & 3.6 \textcolor{red}{(-23.6\%)} & 4.3 \textcolor{red}{(-13.8\%)} \\
    
    \midrule
    \multirow{2}{*}{ViT-L} 
    & LoRA~\cite{hu2021lora} & 76.0 & 8.67 & 61.8 & 19.6\\
    & FPET-LoRA & 76.0 & 6.60 \textcolor{red}{(-23.9\%)} & 46.7 \textcolor{red}{(-24.4\%)} & 15.5 \textcolor{red}{(-21.0\%)} \\
    
    \midrule
    \multirow{4}{*}{DeiT-S} 
    &LoRA\cite{hu2021lora} & 70.1 & 1.08 & 4.8 & 5.0 \\
    &FPET-LoRA & 70.0 & 0.93\textcolor{red}{(-14.2\%)} & 3.6\textcolor{red}{(-23.5\%)} & 4.3\textcolor{red}{(-13.9\%)} \\
    & Bi-LoRA~\cite{jie2023revisiting} & 70.2 & 1.08 & 4.8 & 5.0 \\
    & FPET-Bi-LoRA & 70.2 & 0.93 \textcolor{red}{(-14.2\%)} & 3.6 \textcolor{red}{(-23.5\%)} & 4.3 \textcolor{red}{(-13.9\%)} \\ 

    \midrule
    \multirow{4}{*}{DeiT-B} 
    & LoRA~\cite{hu2021lora} & 72.9 & 2.62 & 17.6 & 8.4 \\
    & FPET-LoRA & 72.8 & 2.10 \textcolor{red}{(-18.7\%)} & 13.3 \textcolor{red}{(-24.4\%)} & 7.1 \textcolor{red}{(-15.5\%)} \\ 
    &AdaptFormer\cite{chen2022adaptformer} & 72.7 & 2.73 & 17.7 & 7.7 \\
    &FPET-AdaptFormer & 72.6 & 2.15\textcolor{red}{(-21.9\%)} & 13.5\textcolor{red}{(-21.9\%)} & 6.2\textcolor{red}{(-21.9\%)} \\

    \bottomrule
  \end{tabular}}
  \caption{Model performance on other backbones.}
  \label{tab:otherbackbone}
\end{table}

\begin{table}[t] 
  \centering
  \resizebox{\linewidth}{!}{
  \begin{tabular}{@{}lcccc@{}}
    \toprule
     Method & Acc (\%) & Time (ms) & FLOPs (G) & Mem (GB) \\
    \midrule
    AdaptFormer (dim=8) & 76.2 & 2.68 & 17.61 & 7.64\\
    AdaptFormer (dim=1) & 74.7 & 2.64 & 17.59 & 7.62\\
    FPET-AdaptFormer & 76.2 & 2.12 & 13.45 & 6.21 \\
    \bottomrule
  \end{tabular}}
  \caption{Comparison between reducing rank redundancy and token redundancy.}
  \label{tab:rank}
\end{table}

\subsection{Trade-offs between efficiency and accuracy}
We provide numerical data corresponding to Fig 3. of the main paper in \cref{tab:layer}. 
As shown in \cref{tab:layer}, our original implementation, which operates at layer 6, achieves a 20.9\% reduction in inference time without compromising accuracy compared to the original AdaptFormer~\cite{chen2022adaptformer} implementation. Furthermore, greater efficiency gains can be achieved by applying our module to earlier layers.

\begin{figure}[!t]
     \centering
     \includegraphics[width=\linewidth]{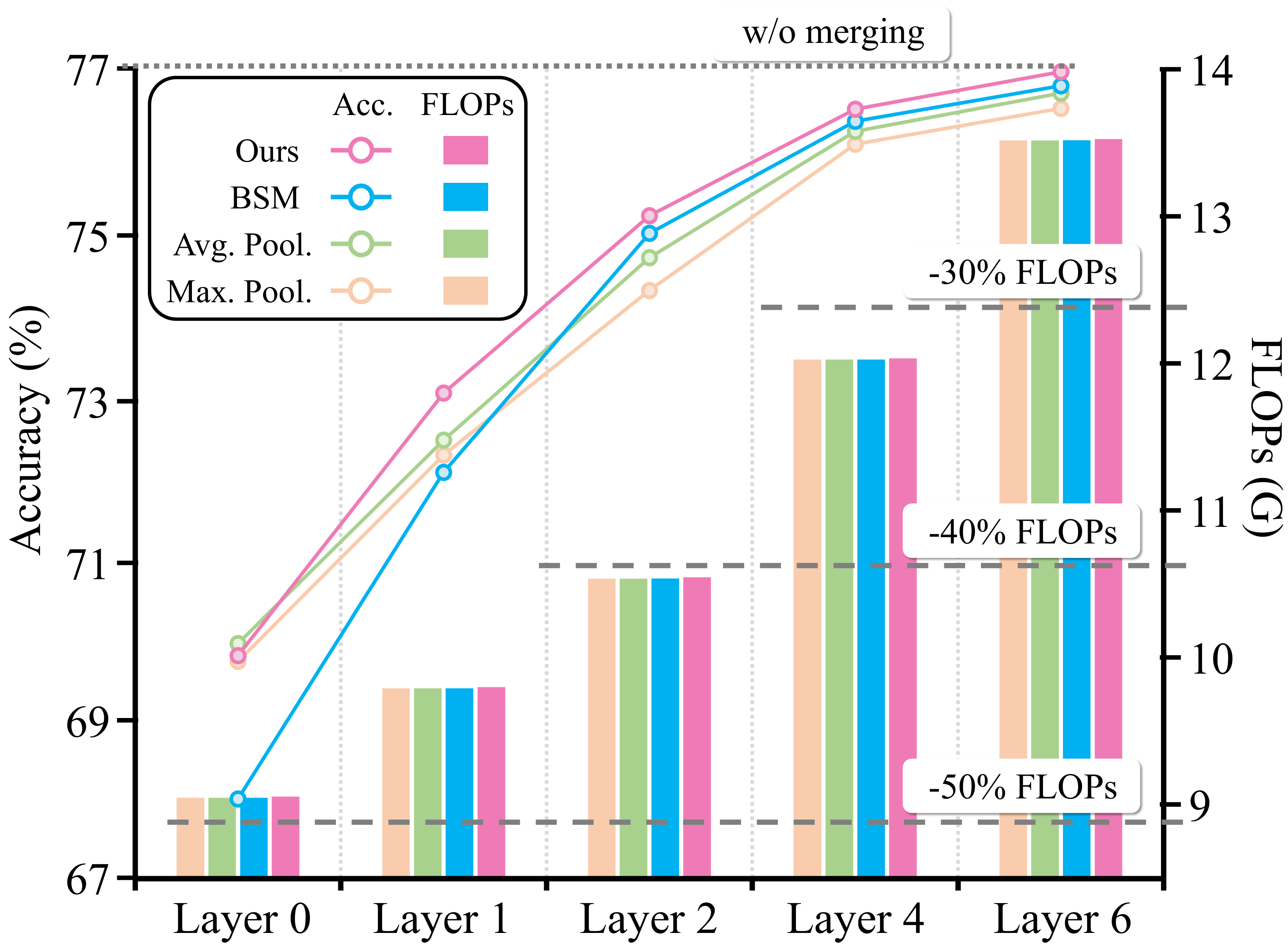}
     \hfill
     \vspace{-10pt}
     \caption{Trade-off between flops and accuracy. All models are ViT-B/16 consisting of 12 transformer layers with Bi-AdaptFormer~\cite{jie2023revisiting}. BSM refers to bipartite soft matching~\cite{bolya2022token}. All methods reduce 196 tokens to 98 tokens at different layers.}
    \label{fig:layer_flops_bi}
\end{figure}

We present the same experimental results on Bi-AdaptFormer~\cite{jie2023revisiting} in \cref{tab:layer_bi} and \cref{fig:layer_flops_bi}. Compared to the original Bi-AdaptFormer, our implementation at layer 6 achieves a 23.3\% reduction in FLOPs. By applying our module at layer 4, we achieve a 31.8\% reduction in FLOPs while maintaining competitive accuracy relative to state-of-the-art PET and efficiency-focused PET methods, as shown in Tab. 1 of the main paper. At layer 2, we further reduce FLOPs by 40.6\%, demonstrating the scalability and efficiency of our approach.

Compared to the bipartite soft matching~\cite{bolya2022token}, our method consistently demonstrates higher accuracy, highlighting the more optimal matching achieved by our approach. This is particularly evident in early layers, where the refinement of similarity is crucial. In these layers, the accuracy of bipartite soft matching~\cite{bolya2022token} drops significantly, performing even worse than basic pooling methods, further underscoring the effectiveness of our method.

\begin{table}

  \centering
  \resizebox{\linewidth}{!}{
  \begin{tabular}{@{}c|lccc@{}}
    \toprule
     Layer & Method & Acc (\%) & Time (ms) & FLOPs (G)\\

    \midrule
    \multirow{ 1}{*}{N/A}
    & w/o merging & 76.22 & 2.68 & 17.6\\

    \midrule
    \multirow{ 4}{*}{6}
    & Max Pool.                         & 75.68 & 2.06  & 13.5\\
    & Avg. Pool.                        & 75.87 & 2.08  & 13.5\\
    & Bipartite soft matching~\cite{bolya2022token}           & 76.03 & 2.12  & 13.5\\ 
    & Bipartite differentiable matching & 76.22 & 2.12  & 13.5\\ 

    \midrule
    \multirow{ 4}{*}{4}
    & Max Pool.                         & 75.03 & 1.89  & 12.0\\
    & Avg. Pool.                        & 75.13 & 1.89  & 12.0\\
    & Bipartite soft matching~\cite{bolya2022token}           & 75.40 & 1.92  & 12.0\\ 
    & Bipartite differentiable matching & 75.65 & 1.92  & 12.0\\ 

    \midrule
    \multirow{ 4}{*}{2}
    & Max Pool.                         & 73.41 & 1.67  & 10.5\\
    & Avg. Pool.                        & 73.43 & 1.68  & 10.5\\
    & Bipartite soft matching~\cite{bolya2022token}           & 73.95 & 1.71  & 10.5\\ 
    & Bipartite differentiable matching & 74.29 & 1.71  & 10.5\\ 

    \midrule
    \multirow{ 4}{*}{1}
    & Max Pool.                         & 71.46 & 1.56  & 9.8\\
    & Avg. Pool.                        & 71.45 & 1.57  & 9.8\\
    & Bipartite soft matching~\cite{bolya2022token}           & 71.06 & 1.58  & 9.8\\ 
    & Bipartite differentiable matching & 72.21 & 1.59  & 9.8\\ 
    
    \midrule
    \multirow{ 4}{*}{0}
    & Max Pool.                         & 68.95 & 1.40  & 9.0\\
    & Avg. Pool.                        & 69.33 & 1.42  & 9.0\\
    & Bipartite soft matching~\cite{bolya2022token}          & 67.39 & 1.44  & 9.0\\ 
    & Bipartite differentiable matching & 69.09 & 1.45  & 9.0\\ 
      
    \bottomrule
  \end{tabular}}
  \caption{Trade-off between inference time and accuracy. All models are ViT-B/16 consisting of 12 transformer layers with AdaptFormer~\cite{chen2022adaptformer}. All methods reduce 196 tokens to 98 tokens at different layers.}
  \label{tab:layer}
  \vspace{-5pt}
\end{table}

\begin{table}

  \centering
  \resizebox{\linewidth}{!}{
  \begin{tabular}{@{}c|lccc@{}}
    \toprule
     Layer & Method & Acc (\%) & Time (ms) & FLOPs (G)\\

    \midrule
    \multirow{ 1}{*}{N/A}
    & w/o merging & 77.01 & 2.77 & 17.7\\

    \midrule
    \multirow{ 4}{*}{6}
    & Max Pool.                         & 76.65 & 2.13  & 13.5\\
    & Avg. Pool.                        & 76.79 & 2.14  & 13.5\\
    & Bipartite soft matching~\cite{bolya2022token}           & 76.83 & 2.17  & 13.5\\ 
    & Bipartite differentiable matching & 76.96 & 2.17  & 13.5\\ 

    \midrule
    \multirow{ 4}{*}{4}
    & Max Pool.                         & 76.06 & 1.92  & 12.0\\
    & Avg. Pool.                        & 76.22 & 1.91  & 12.0\\
    & Bipartite soft matching~\cite{bolya2022token}           & 76.50 & 1.94  & 12.0\\ 
    & Bipartite differentiable matching & 76.34 & 1.94  & 12.0\\ 

    \midrule
    \multirow{ 4}{*}{2}
    & Max Pool.                         & 74.26 & 1.71  & 10.5\\
    & Avg. Pool.                        & 74.66 & 1.71  & 10.5\\
    & Bipartite soft matching~\cite{bolya2022token}           & 74.96 & 1.72  & 10.5\\ 
    & Bipartite differentiable matching & 75.18 & 1.71  & 10.5\\ 

    \midrule
    \multirow{ 4}{*}{1}
    & Max Pool.                         & 72.23 & 1.58  & 9.8\\
    & Avg. Pool.                        & 72.41 & 1.59  & 9.8\\
    & Bipartite soft matching~\cite{bolya2022token}           & 72.02 & 1.60  & 9.8\\ 
    & Bipartite differentiable matching & 73.00 & 1.60  & 9.8\\ 
    
    \midrule
    \multirow{ 4}{*}{0}
    & Max Pool.                         & 69.68 & 1.51  & 9.0\\
    & Avg. Pool.                        & 70.31 & 1.52  & 9.0\\
    & Bipartite soft matching~\cite{bolya2022token}          & 67.98 & 1.54  & 9.0\\ 
    & Bipartite differentiable matching & 69.75 & 1.53  & 9.0\\ 
      
    \bottomrule
  \end{tabular}}

  \caption{Trade-off between inference time and accuracy. All models are ViT-B/16 consisting of 12 transformer layers with Bi-AdaptFormer~\cite{jie2023revisiting}. All methods reduce 196 tokens to 98 tokens at different layers.}
  \label{tab:layer_bi}
  \vspace{-7pt}
\end{table}

\subsection{Performance on full datasets}
\label{sec:full}

In line with previous studies \cite{jia2022visual, jie2023fact, jie2023revisiting, lian2022scaling, zhang2022neural}, our models are trained on VTAB-1K~\cite{zhai2019visual} training set which comprises subset of each original downstream dataset. We extend the evaluation of our method to training on full datasets of CIFAR100 \cite{cifar} and SVHN \cite{svhn}, which contain 50,000 and 73,257 training images, respectively. 

As outlined in the main paper, our implementation utilizes the ViT-B/16 as the backbone model, with the hidden dimension set to 8 for both AdaptFormer \cite{chen2022adaptformer} and LoRA \cite{hu2021lora}.
As shown in ~\cref{tab:full}, our method achieves competitive accuracy relative to the original implementations underscoring the robustness of FPET in full dataset scenarios.

\begin{table}[]
  \centering
  \begin{tabular}{@{}lcc@{}}
    \toprule
     Method & CIFAR-100 & SVHN \\
    \midrule
    AdaptFormer & 92.14 & 97.21 \\
    FPET-AdaptFormer & 92.18 & 97.18 \\
    LoRA & 92.17 & 97.45 \\
    FPET-LoRA & 92.29 & 97.48 \\
    \bottomrule
  \end{tabular} 
  \caption{Accuracy on full CIFAR-100 and SVHN dataset.}
  \label{tab:full}
\end{table}

\begin{figure}[t!]
\centering
\includegraphics[width=\linewidth]{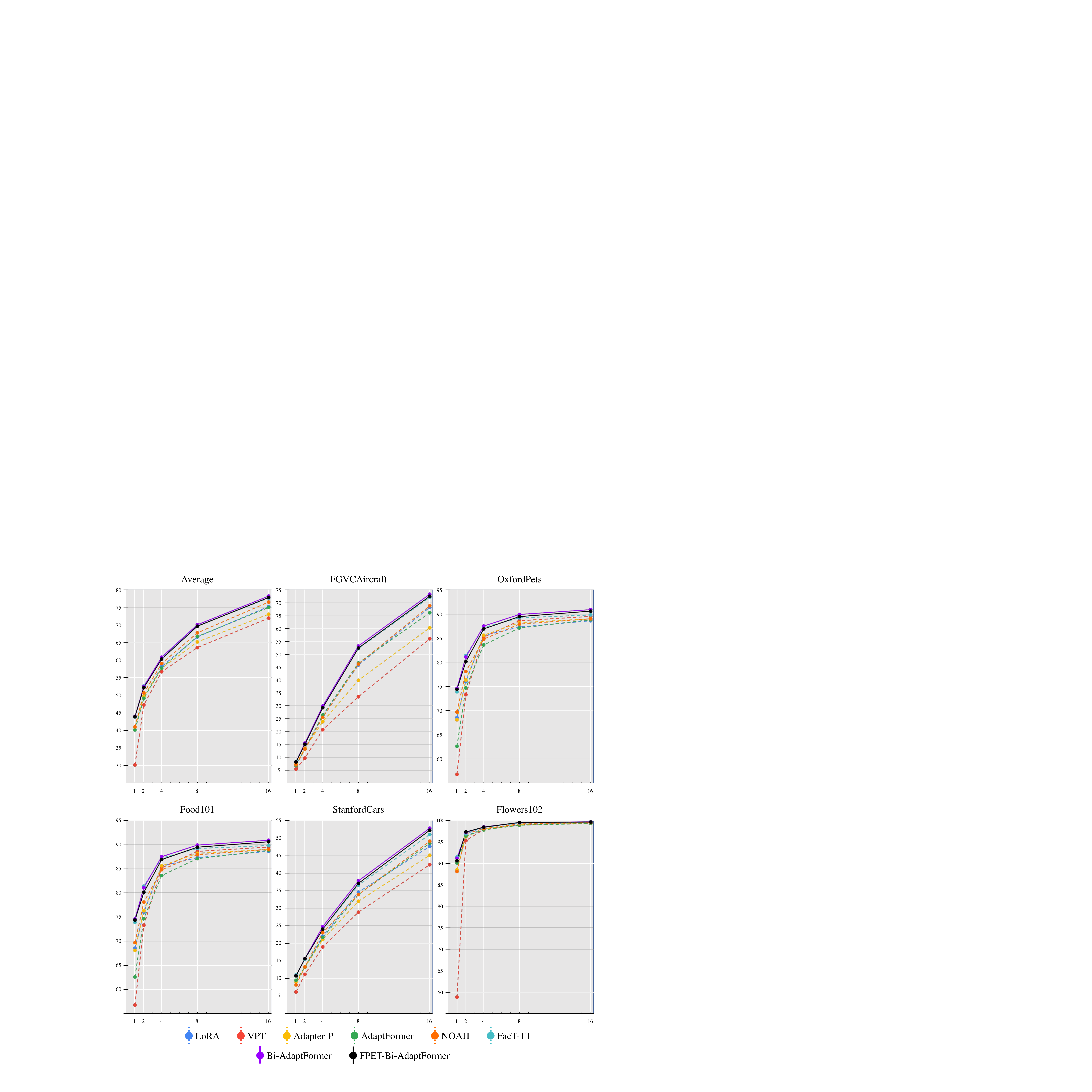}
\caption{Performance of Few-shot learning on FGVC dataset including FGVC-Aircraft~\cite{maji2013fine}, Oxford-Pets~\cite{pets}, Food-101~\cite{food101}, Stanford Cars~\cite{cars} and Oxford-Flowers102~\cite{flowers}.}
\label{fig:fig_fgvc}
\end{figure}

\begin{figure*}[t!]
     \centering
     \begin{subfigure}[b]{\textwidth}
         \centering
         \includegraphics[width=\textwidth]{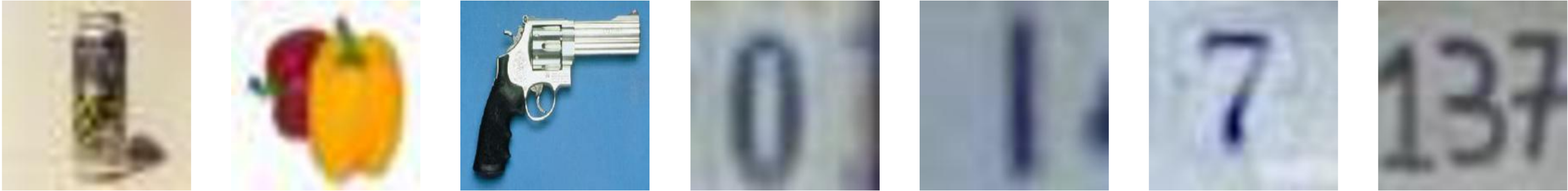}
         \caption{Input images}
         \label{fig:fig_vis_a}
     \end{subfigure}
     \begin{subfigure}[b]{\textwidth}
         \centering
         \includegraphics[width=\textwidth]{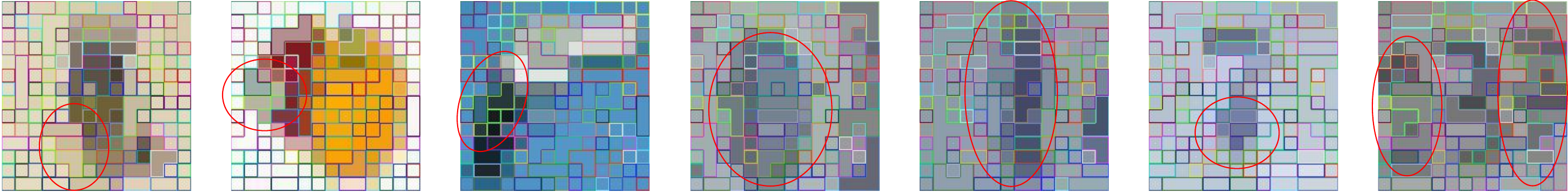}
         \caption{ToMe~\cite{bolya2022token}}
         \label{fig:fig_vis_d}
     \end{subfigure}
     \begin{subfigure}[b]{\textwidth}
         \centering
         \includegraphics[width=\textwidth]{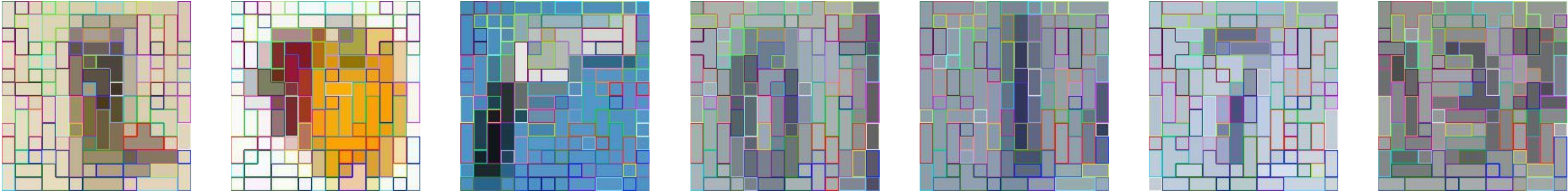}
         \caption{FPET}
         \label{fig:fig_vis_b}
     \end{subfigure}
     \caption{Token merging visualization. We visualize input patches associated with each merged token at the end of networks, following the methodology provided by \cite{bolya2022token}. Patches belonging to the same merging group are represented by the same averaged color and bordered by a color randomly assigned to each group. Blurred boundaries and distortions indicating semantically irrelevant merging are highlighted in red circles. We employ ViT-B/16 as the backbone and Bi-AdaptFormer as the PET method.}
    \label{fig:fig_vis}
\end{figure*}

\subsection{Visualization}

In \cref{fig:fig_vis}, we visualize token merging results produced by ToMe~\cite{bolya2022token}and our token redundancy reduction module to demonstrate the effectiveness of the proposed method.
Employing the visualization techniques in \cite{bolya2022token}, we trace the trajectory of each token and project the merging results onto the original image by averaging the colors within each group. This method ensures that patches belonging to the same merging group are represented by the same averaged color and bordered by a color randomly assigned to each group, facilitating a clear visual distinction.

Notably, whereas ToMe merges 8 tokens across all 12 layers, resulting in a total of 96 tokens to be merged, FPET executes a one-time merge of 98 tokens, equivalent to half of the total patch count, at the middle layer only. 
Despite the larger number of remaining tokens in the last layer, the visual evidence in \cref{fig:fig_vis} reveals that ToMe\cite{bolya2022token} often produces images with blurred boundaries and distortions as highlighted in red circles. These results indicate semantically irrelevant merges between the object patches and background or other object patches. In contrast, FPET maintains sharper edges and more authentic shapes, demonstrating more optimal merging outcomes to preserve crucial visual information. This comparison underscores our module delicately engineered for PET and the enhanced precision and effectiveness of the merging process in FPET.

\subsection{Accuracy Drop of Other Token Reduction Methods}
We report the accuracy drop relative to the original implementations of various token reduction methods across different PET methods. As shown in \cref{tab:tokenreduction_pet}, our method consistently achieves smaller accuracy drops compared to existing approaches.

\begin{table}[t] 
  \centering
  \begin{tabular}{@{}lccc@{}}
    \toprule
     Method & LTMP~\cite{bonnaerens2023learned} & ToMe~\cite{bolya2022token} & Ours \\
    \midrule
    RepAdapter & -0.32 & -0.33 & \textbf{-0.05}\\
    LoRA~\cite{hu2021lora} & -0.32 & -0.32 & \textbf{-0.12} \\
    AdaptFormer~\cite{chen2022adaptformer} & -0.58 & -0.34 & \textbf{0.00} \\
    Bi-LoRA~\cite{jie2023revisiting} & -0.34 & -0.34 & \textbf{-0.17} \\
    Bi-AdaptFormer~\cite{jie2023revisiting} & -0.38 & -0.27 & \textbf{-0.05} \\
    \bottomrule
  \end{tabular}
  \caption{Accuracy drop on other token reduction methods. All models are ViT-B/16 consist of 12 transformer layers with respective PET method. Each model starts with 197 tokens. LTMP reduces a variable number of tokens, while ToMe merges 8 tokens at each of the 12 layers. }

  \label{tab:tokenreduction_pet}
  
 \end{table}

\section{Performance on Other Tasks}
\subsection{Performance on Few-shot learning}
\label{sec:fewshot}

To assess performance of FPET in low-data scenarios, we conduct experiments using the FGVC dataset. Following the setting in \cite{jie2023revisiting}, we utilize five datasets, FGVC-Aircraft~\cite{maji2013fine}, Oxford-Pets~\cite{pets}, Food-101~\cite{food101}, Stanford Cars~\cite{cars}, and Oxford-Flowers102~\cite{flowers}. Our experiments span 1, 2, 4, 8, and 16-shot settings. 

We implement FPET on Bi-AdaptFormer\cite{jie2023revisiting} and compare our Bi-AdaptFormer-FPET with several state-of-the-art approaches, including LoRA\cite{hu2021lora}, VPT\cite{jia2022visual}, Adapter-P\cite{pfeiffer2020adapterfusion}, AdaptFormer\cite{chen2022adaptformer}, NOAH\cite{zhang2022neural}, FacT-TT\cite{jie2023fact}, and Bi-AdaptFormer\cite{jie2023revisiting}.
As in \cite{jie2023revisiting}, we configure the hidden dimension as 8 for AdaptFormer \cite{chen2022adaptformer}, LoRA \cite{hu2021lora}, and Adapter-P\cite{pfeiffer2020adapterfusion}, and as 32 for Bi-AdaptFormer \cite{jie2023revisiting} and Bi-AdaptFormer-FPET. The prompt length for VPT is set to 8, while the rank for FacT-TT\cite{jie2023fact} is determined to be 16. For NOAH\cite{zhang2022neural}, we adopt the best configuration as suggested in their paper.

As depicted in \cref{fig:fig_fgvc}, Bi-AdaptFormer \cite{jie2023revisiting} emerges as the top performer regarding accuracy. Our approach, Bi-AdaptFormer-FPET, demonstrates competitive accuracy in comparison, thereby underscoring the robustness of FPET in scenarios characterized by limited data availability.

\subsection{Performance on Cross-modal Retrieval}
We present the cross-modal retrieval performance evaluated on the Flickr30K dataset~\cite{plummer2015flickr30k}, to further demonstrate the scalability of our method on vision-language tasks. Specifically, we employ the pretrained BLIP-base~\cite{li2022blip} as our vision-language backbone and apply the proposed FPET to the vision model. 
As most of the latency arises from the vision model, applying our method to only the pre-trained vision model sufficiently yields significant efficiency improvements. 
As a baseline, we compare the results with UniAdapter~\cite{lu2023uniadapter} that shares the same spirit of our method.
As shown in \cref{tab:retrieval}, we achieve a 28.44\% reduction in training time and a 22.62\% reduction in GPU memory usage, demonstrating the practicality of our approach in resource-constrained environments. 

\begin{table*}[t] 
  \centering
  \resizebox{\linewidth}{!}{
  \begin{tabular}{l ccc ccc cc}
    \toprule
     \multirow{2}[2]{*}{Method} &  \multicolumn{3}{c}{I2T Retrieval} & \multicolumn{3}{c}{T2I Retrieval} & \multicolumn{2}{c}{Efficiency} \\
     \cmidrule(lr){2-4} \cmidrule(lr){5-7} \cmidrule(lr){8-9}
      & R@1 & R@5 & R@10 & R@1 & R@5 & R@10 & Training time (hr/epoch) & Mem (GB) \\
    \midrule
    UniAdapter~\cite{lu2023uniadapter} & 94.2 & 99.5 & 99.7 & 83.6 & 96.6 & 98.2  & 1.09 & 44.96 \\
    FPET-UniAdapter & 94.1 & 99.4 & 99.9 & 83.0 & 96.0 & 98.0 & 0.78\textcolor{red}{(-28.44\%)}	& 34.79\textcolor{red}{(-22.62\%)} \\
    \bottomrule
  \end{tabular}}
  \caption{Performance on cross-modal retrieval evaluated on Flickr30K~\cite{plummer2015flickr30k}. We report R@1, 5, and 10 for image-to-text (I2T) and text-to-image (T2I) retrieval, and efficiency in terms of training time and memory.}
  \label{tab:retrieval}
\end{table*}

\section{Pseudo code}
\label{sec:alg}
To succinctly present the implementation of our token redundancy reduction module, we provide its pseudo code representation in \cref{fig:code}. 

\begin{figure*}[!t]
  \centering

\begin{python}
import torch

# bipartite differentiable matching
def bdm(self, tokens, key):
    
    # halt propagation of gradients to the backbone
    key = key.detach()
    
    # key refinement
    key_refined = key + self.refinement(key)

    # split tokens
    k_a, k_b = checkerboard_split(key_refined)
    x_a, x_b = checkerboard_split(tokens)
    
    # refined similarity matrix
    scores = k_a @ k_b.transpose(-1, -2)
    
    # exclude cls token
    scores[..., :, 0] = -math.inf 
    
    # average of the top-1 and top-2 values
    v, idx = torch.topk(scores, 2, dim=-1)
    mean12 = v.mean(dim=-1, keepdim=True)

    # generate the soft matching matrix
    soft_matrix = torch.sigmoid(scores - mean12)
    
    # generate the hard matching matrix
    hard_matrix = (soft_mask > 0.5).float()
    
    # generate the matching matrix
    matching_matrix = soft_matrix + (hard_matrix - soft_matrix).detach()

    # merging tokens
    x_merged_sum = x_b + torch.einsum('bik, bij->bkj', matching_matrix, x_a)
    self.size = self.size_update(self.size, matching_matrix)
    x_merged = self.average(x_merged_sum, self.size)

    return x_merged
\end{python}
\caption{Implementation our proposed bipartite differentiable matching.}
\label{fig:code}
\end{figure*}

\section{Hyper-parameters}
\label{sec:hyper}

We provide the hyper-parameters for our experiments in \cref{tab:hyper}.

\begin{table*}
  \centering
  \begin{tabular}{@{}ccccccc@{}}
    \toprule
    optimizer & batch size & learning rate & weight decay & \# epochs & lr decay & \# warm-up epochs \\
    \midrule
    AdamW & 64 & 1e-3 & 1e-4 & 100 & cosine & 10 \\
    \bottomrule
  \end{tabular} 
  \label{tab:hyper}
  \caption{Hyper-parameters for our experiments.}
\end{table*}

\section{Datasets}
\label{sec:datasets}

We provide the statistics of datasets used for our experiments in \cref{tab:datasets}.

\begin{table*}[t!]
    \centering
    \begin{tabular}{@{}clcccc@{}}
    \toprule
    & Dataset & \# Classes & Train & Val & Test \\ 
    \midrule
    \multicolumn{6}{@{} c}{VTAB-1K~\cite{zhai2019visual}} \\
    \midrule
    \multirow{7}{*}{Natural}
    & CIFAR100~\cite{cifar} & 100 & \multirow{7}{*}{800/1,000} & \multirow{7}{*}{200} & 10,000 \\
    & Caltech101~\cite{caltech} & 102 & & & 6,084 \\
    & DTD~\cite{cimpoi2014describing} & 47 & & & 1,880 \\
    & Oxford-Flowers102~\cite{flowers} & 102 & & & 6,149 \\
    & Oxford-Pets~\cite{pets} & 37 & & & 3,669 \\
    & SVHN~\cite{svhn} & 10 & & & 26,032 \\
    & Sun397~\cite{sun} & 397 & & & 21,750 \\
    \midrule
    \multirow{4}{*}{Specialized}
    & Patch Camelyon~\cite{veeling2018rotation} & 2 & \multirow{4}{*}{800/1,000} & \multirow{4}{*}{200} & 32,768 \\
    & EuroSAT~\cite{helber2019eurosat} & 10 &  & & 5,400 \\
    & Resisc45~\cite{cheng2017remote} & 45 &  & & 6,300 \\
    & Retinopathy~\cite{retinopathy} & 5 &  & & 42,670 \\
    \midrule
    \multirow{8}{*}{Structured}
    & Clever/count~\cite{johnson2017clevr} & 8 & \multirow{8}{*}{800/1,000} & \multirow{8}{*}{200} & 15,000 \\
    & Clever/distance~\cite{johnson2017clevr} & 6 & & & 15,000 \\
    & DMLab~\cite{beattie2016deepmind} & 6 & & & 22,735 \\
    & KITTI-Dist~\cite{geiger2013vision} & 4 & & & 711 \\
    & dSperites/location~\cite{higgins2016beta} & 16 & & & 73,728 \\
    & dSperites/orientation~\cite{higgins2016beta} & 16 & & & 73,728 \\
    & SmallNORB/azimuth~\cite{smallnorb} & 18 & & & 12,150 \\
    & SmallNORB/elevation~\cite{smallnorb} & 18 & & & 12,150 \\
    \midrule
    \multicolumn{6}{@{} c}{Few-shot learning} \\
    \midrule
    \multirow{8}{*}{}
    & Food-101 \cite{food101} & 101 & \multirow{5}{*}{1/2/4/8/16 per class} & 20,200 & 30,300 \\
    & Stanford Cars\cite{cars} & 196 & & 1,635 & 8,041 \\
    & Oxford-Flowers102\cite{flowers} & 102 & & 1,633 & 2,463 \\
    & FGVC-Aircraft\cite{maji2013fine} & 100 & & 3,333 & 3,333 \\
    & Oxford-Pets\cite{pets} & 37 & & 736 & 3,699 \\
    \midrule
    \multicolumn{6}{@{} c}{Full datasets} \\
    \midrule
    & CIFAR100\cite{cifar} & 100 & 60,000 & - & 10,000 \\
    & SVHN\cite{svhn} & 10 & 73,257 & - & 26,032 \\
    \bottomrule
    \end{tabular}
    \caption{Statistics of datasets.}
    \label{tab:datasets}
\end{table*}

\clearpage
\clearpage


\end{document}